\documentclass[letterpaper]{article} 
\usepackage{aaai23}  

\usepackage{xr}

\usepackage{times}  
\usepackage{helvet}  
\usepackage{courier}  
\usepackage[hyphens]{url}  
\usepackage{graphicx} 
\usepackage{natbib}  
\usepackage{caption} 
\usepackage{algorithm}
\usepackage{algpseudocode}

\usepackage{newfloat}
\usepackage{listings}
\DeclareCaptionStyle{ruled}{labelfont=normalfont,labelsep=colon,strut=off} 
\floatstyle{ruled}
\newfloat{listing}{tb}{lst}{}
\floatname{listing}{Listing}
\usepackage[dvipsnames]{xcolor}
\usepackage{amsmath,amsthm,amssymb,amsfonts}
\usepackage{todonotes}
\usepackage{mathrsfs}
\usepackage{subcaption}
\usepackage{dblfloatfix}

\DeclareMathOperator*{\argmin}{arg min}

\DeclareMathOperator{\Ex}{\mathbb{E}}
\newcommand{\R}{\mathbb{R}}
\newcommand{\s}{\mathcal{S}}
\newcommand{\A}{\mathcal{A}}
\newcommand{\M}{\mathcal{M}}
\newcommand{\T}{\mathcal{T}}
\newcommand{\B}{\mathcal{B}}
\newcommand{\D}{\mathcal{D}}
\newcommand{\loss}{\mathcal{L}}
\newcommand{\E}{\mathbb{E}}

\theoremstyle{definition}
\newtheorem{definition}{Definition}

\newtheorem{assumption}{Assumption}

\newcommand{\charlotte}[1]{\todo[size=\small, color=Purple]{Charlotte}}

\newcommand{\francois}[1]{\todo[size=\small, color=Blue]{Francois}}

\makeatletter
\newcommand*{\addFileDependency}[1]{
  \typeout{(#1)}
  \@addtofilelist{#1}
  \IfFileExists{#1}{}{\typeout{No file #1.}}
}
\makeatother

\newcommand*{\myexternaldocument}[1]{
    \externaldocument{#1}
    \addFileDependency{#1.tex}
    \addFileDependency{#1.aux}
}
\myexternaldocument{supp_material}

\title{Hypernetworks for Zero-shot Transfer in Reinforcement Learning}
\author{
    Sahand Rezaei-Shoshtari\textsuperscript{\rm 1,2,3}, Charlotte Morissette\textsuperscript{\rm 1,3}, Francois R. Hogan\textsuperscript{\rm 3}\\ Gregory Dudek\textsuperscript{\rm 1,2,3}, David Meger\textsuperscript{\rm 1,2,3}
}
\affiliations{
    \textsuperscript{\rm 1}McGill University
    \textsuperscript{\rm 2}Mila - Québec AI Institute
    \textsuperscript{\rm 3}Samsung AI Center Montreal

    srezaei@cim.mcgill.ca
}

\begin{document}

\maketitle

\begin{abstract}
In this paper, hypernetworks are trained to generate behaviors across a range of unseen task conditions, via a novel TD-based training objective and data from a set of near-optimal RL solutions for training tasks. This work relates to meta RL, contextual RL, and transfer learning, with a particular focus on  zero-shot performance at test time, enabled by knowledge of the task parameters (also known as context). Our technical approach is based upon viewing each RL algorithm as a mapping from the MDP specifics to the near-optimal value function and policy and seek to approximate it with a hypernetwork that can generate near-optimal value functions and policies, given the parameters of the MDP. We show that, under certain conditions, this mapping can be considered as a supervised learning problem. We empirically evaluate the effectiveness of our method for zero-shot transfer to new reward and transition dynamics on a series of continuous control tasks from DeepMind Control Suite. Our method demonstrates significant improvements over baselines from multitask and meta RL approaches. 
\end{abstract}

\section{Introduction}
\label{sec:intro}
Adult humans possess an astonishing ability to adapt their behavior to new situations. Well beyond simple tuning, we can adopt entirely novel ways of moving our bodies, for example walking on crutches with little to no training after an injury. The learning process that generalizes across all past experience and modes of behavior to rapidly output the needed behavior policy for a new situation is a hallmark of our intelligence. 

\begin{figure}[t!]
    \centering
    \includegraphics[width=0.49\textwidth]{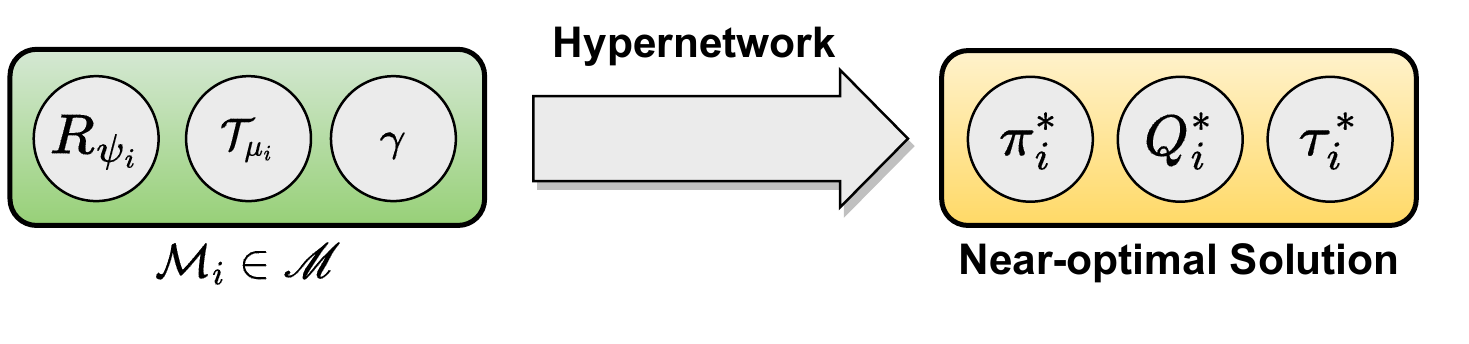}
    \caption{Our method uses hypernetworks to approximate an RL algorithm as a mapping from a family of parameterized MDPs to a family of near-optimal solutions, in order to achieve zero-shot transfer to new reward and dynamics settings.}
    \label{fig:urla_diag}
\end{figure}

This paper proposes a strong zero-shot behavior generalization approach based on hypernetworks~\cite{ha2016hypernetworks}, a recently proposed architecture allowing a deep hyper-learner to output all parameters of a target neural network, as depicted in Figure \ref{fig:urla_diag}. In our case, we train on the full solutions of numerous RL problems in a family of MDPs, where either reward or dynamics (often both) can change between task instances. The trained policies, value functions and rolled-out optimal behavior of each source task is the training information from which we can learn to generalize. Our hypernetworks output the parameters of a fully-formed and highly performing policy without any experience in a related but unseen task, simply by conditioning on provided task parameters.  

The differences between the tasks we consider leads to large and complicated changes in the optimal policy and induced optimal trajectory distribution. Learning to predict new policies from this data requires powerful learners guided by helpful loss functions. We show that the abstraction and modularity properties afforded by hypernetworks allow them to approximate RL generated solutions by mapping a parameterized MDP family to a set of optimal solutions. We show that this framework enables achieving strong zero-shot transfer to new reward and dynamics settings by exploiting  commonalities in the MDP structure.  

We perform experimental validation using several families of continuous control environments where we have parameterized the physical dynamics, the task reward, or both to evaluate learners. We carry out contextual zero-shot evaluation, where the learner is provided the parameters of the test task, but is not given any training time -- rather the very first policy execution at test time is used to measure performance. Our method outperforms selected well-known baselines, in many cases recovering nearly full performance without a single timestep of training data on the target tasks. Ablations show that hypernetworks are a critical element in achieving strong generalization and that a structured TD-like loss is additionally helpful in training these networks.

Our main contributions are: 
\begin{enumerate}
    \item The use of hypernetworks as a scalable and practical approach for approximating RL algorithms as a mapping from a family of parameterized MDPs to a family of near-optimal policies. 
    \item A TD-based loss for regularization of the generated policies and value functions to be consistent with respect to the Bellman equation.
    \item A series of modular and customizable continuous control environments for transfer learning across different reward and dynamics parameters.
\end{enumerate}
Our learning code, generated datasets, and custom continuous control environments, which are built upon DeepMind Control Suite, are publicly available at: {\footnotesize\texttt{https://sites.google.com/view/hyperzero-rl}}

\section{Background}
\label{sec:bg}
\subsection{Markov Decision Processes}
\label{sec:bg_mdp}
We consider the standard MDP that is defined by a 5-tuple $\M = (\s, \A, \T, R, \gamma)$, where  $\s$ is the state space,  $\A$ is the action space, $\T: \s \times \A \to \text{Dist}({\s})$ is the transition dynamics, $R : \s \times \A \to \mathbb{R}$ is the reward function and $\gamma \in (0, 1]$ is the discount factor. The goal of an RL algorithm is to find a policy $\pi : \s \to \text{Dist}(\A)$ that maximizes the \emph{expected return} defined as $\Ex_\pi [R_t] = \Ex [\sum_{k=0}^T \gamma^k r_{t+k+1}]$. \emph{Value function} $V^\pi(s)$ denotes the expected return from $s$ under policy $\pi$, and similarly $\emph{action-value function}$ $Q^\pi(s, a)$ denotes the expected return from $s$ after taking action $a$ under policy $\pi$:
$$
Q^\pi(s, a) = \mathbb{E}_{s', r \sim p(\cdot|s, a), a' \sim \pi(\cdot | s')} \Big[\sum_{k=0}^\infty \gamma^k r_{t+k+1} \Big| s, a \Big]
$$
Value functions are fixed points of the Bellman equation \cite{bellman1966dynamic}, or equivalently the Bellman operator $\B^\pi$:
$$
    \B^\pi [Q(s, a)] = \E_{s', r \sim p(\cdot|s, a), a' \sim \pi(\cdot | s')} \Big[r + \gamma Q(s', a') \Big]
$$
Similarly, the optimal value functions $V^*(s)$ and $Q^*(s, a)$ are the fixed points of the Bellman optimality operator $\B^*$.

\subsection{General Value Functions}
\label{sec:bg_gvf}
General value functions (GVF) extend the standard definition of value functions $Q^\pi (s, a)$ to entail the reward function, transition dynamics and discount factor in addition to the policy, that is $Q^{\pi, R, \T, \gamma}(s, a)$ \cite{sutton2018reinforcement}. Universal value function approximators (UVFA) \cite{schaul2015universal} are an instance of GVFs in which the value function is generalized across goals $g$ and is represented as $Q^\pi(s, a, g)$. Naturally, this notion is used in goal-conditioned RL \cite{andrychowicz2017hindsight} and multi-task RL \cite{teh2017distral}. Relatedly, general policy improvement (GPI) aims to improve a generalized policy based on transitions of several MDPs \cite{barreto2020fast, harb2020policy, faccio2022general}. The goal of our method in learning a generalized mapping from MDP specifics to near-optimal policies and value functions is closely related to the overall goal of GVFs and GPI. However, unlike such methods we do not seek to improve a given generalized policy.

\subsection{Hypernetworks}
\label{sec:bg_hypernets}
A \emph{hypernetwork} \cite{ha2016hypernetworks} is a neural network that generates the weights of another network, often referred to as the \emph{main} network. While both networks have associated weights, only the hypernetwork weights involve learnable parameters that are updated during training. During inference, only the main network is used by mapping an input to a desired target, using the weights generated by the hypernetwork. Since the weights of different layers of the main network are generated through a shared learned embedding, hypernetworks can be viewed as a relaxed form of weight sharing across layers. It has been empirically shown that this approach allows for a level of abstraction and modularity of the learning problem \cite{galanti2020modularity, ha2016hypernetworks} which in turn results in a more efficient learning. Notably, hypernetworks can be conditioned on the context vector for conditional generation of the weights of the main network \cite{von2019continual}. Similarly to \citet{von2019continual}, we condition the hypernetwork on the parameters (context) of the MDP to generate the near-optimal policy and value function based on the reward and dynamics parameters.

\section{HyperZero}
\label{sec:method}
The overarching goal of this work is to develop a framework that allows for approximating RL solutions by learning the mapping between the MDP specifics and the near-optimal policy. A reasonable approximation can potentially allow for zero-shot transfer and predicting the general behaviour of an RL agent prior to its training. Beyond the standard premises of zero-shot transfer learning \cite{JMLR:v10:taylor09a, tan2018survey}, a well-approximated mapping of an MDP to near-optimal policies can have applications in reward shaping, task visualization, and environment design. 

\subsection{Problem Formulation}
\label{sec:problem_form}
This section outlines the assumptions and problem formulation used in this paper. First, we define  the \emph{parameterized MDP family} $\mathscr{M}$ as:
\begin{definition}[Parameterized MDP Family]
\label{def:mdp_family}
A \emph{parameterized MDP family} $\mathscr{M}$ is a set of MDPs that share the same state space $\s$, action space $\A$, a parameterized transition dynamics $\T_\mu$, a parameterized reward function $R_\psi$, and a discount factor $\gamma$:
$$
    \mathscr{M} = \{ \M_i | \M_i = (\s, \A, \T_{\mu_i}, R_{\psi_i}, \gamma) \},
$$
where $\psi_i \!\sim\! p(\psi)$ and $\mu_i \!\sim\! p(\mu)$ are parameters of $\M_i$, and are assumed to be sampled from prior distributions.
\end{definition}
Notably, the state space $\mathcal{S}$ and action space $\mathcal{A}$ in our definition can be either discrete or continuous (e.g., an open sub-space of $\R^n)$ spaces. 
Our definition of a parameterized MDP family is related to contextual MDPs \cite{hallak2015contextual, jiang2017contextual}, where the learner has access to the context. 

The key to our approximation is to assume that an RL algorithm, once converged, is a mapping from an  MDP $ \M_i \!\in\! \mathscr{M}$ to a near-optimal policy and the near-optimal action-value function  corresponding to the specific MDP $\M_i$ on which it was trained. With a slight abuse of notation, we denote the near-optimal policy as $\pi_i^*$ and the near-optimal action-value function as $Q_i^*$:
\begin{equation}
    \label{eq:rl_map}
    \M_i \xrightarrow{\text{RL Algorithm}} \pi_i^* (a | s), Q_i^*(s, a).
\end{equation}
Notably, this view has precedent in prior works on learning to shape rewards \cite{sorg2010reward, zheng2018learning, zheng2020can}, meta-gradients in RL \cite{xu2018meta, xu2020meta}, and the operator view of RL algorithms \cite{tang2022operator}. 

Since our goal is to learn the mapping of Equation \eqref{eq:rl_map} from a family of parameterized MDPs that share the similar functional form of parameterized transition dynamics $\T_\mu$ and reward function $R_\psi$, we assume that the MDP $\M_i$ can be fully characterized by its parameters $\psi_i$ and $\mu_i$ and the functional forms of $R_\psi$ and $\T_\mu$; that is $\M_i \equiv \M(\psi_i, \mu_i)$. 
Equation \eqref{eq:rl_map} can then be simplified as:
\begin{equation}
    \label{eq:rl_map_reward}
    \M(\psi_i, \mu_i) \!\xrightarrow{\text{RL Algorithm}}\! \pi^*\!(a | s, \psi_i, \mu_i), Q^*(s, a | \psi_i, \mu_i), 
\end{equation}
where the near-optimal policies and action-value functions are now functions of the reward parameters $\psi_i$ and dynamics parameters $\mu_i$, in addition to their standard inputs. Notably, this formulation is closely related to prior works on goal-conditioned RL \cite{andrychowicz2017hindsight, schroecker2020universal}, and universal value function approximators (UVFA) \cite{schaul2015universal, borsa2018universal}.

Consequently, our problem is formally defined as approximating the mapping shown in Equation \eqref{eq:rl_map_reward} to obtain the approximated near-optimal action-value function $\hat{Q}_\phi(s, a | \psi, \mu)$ and policy $\hat{\pi}_\theta (a | s, \psi, \mu)$ that are parameterized by $\phi$ and $\theta$, respectively. Once such mapping is obtained, one can predict and observe near-optimal trajectories by rolling out the approximated policy $\hat{\pi}_\theta$ without necessarily training the RL solver from scratch:
\begin{equation}
    \hat{\pi}_\theta(a | s, \psi, \mu) \xrightarrow{\text{Policy Rollout in the Environment}} \hat{\tau}(\psi, \mu),
\end{equation}
where $\hat{\tau}(\psi, \mu)$ is the near-optimal trajectory corresponding to the reward parameters $\psi$ and dynamics parameters $\mu$. 

\subsection{Generating Optimal Policies and Optimal Value Functions with Hypernetworks}
\label{sec:method_alg}

\label{sec:sl_of_rl}
Our goal is to approximate the mapping described in Equation \eqref{eq:rl_map_reward}. To that end, we assume having access to a family of near-optimal policies $\pi_i^*$ that were trained independently on instances of $M_i \!\in\! \mathscr{M}$. A dataset of near-optimal trajectories is then collected by rolling out each $\pi_i^*$ on its corresponding MDP $\M_i$. Thus, samples are drawn from the stationary state distribution of the near-optimal policy $d^{\pi^*}(s)$. 

Consequently, the inputs to the learner are tuples of states, reward parameters and dynamics parameters $\langle s, \psi_i, \mu_i \rangle$ and the targets are tuples of near-optimal actions and action-values $\langle a^*, q^* \rangle$. 
We can frame the approximation problem of Equation \eqref{eq:rl_map_reward} as a supervised learning problem under the following conditions: 
\begin{assumption}
\label{ass_1} The parameters of the reward function $R_\psi$ and transition dynamics $\T_\mu$ are sampled independently and identically from distributions over the parameters $\psi_i \sim p(\psi)$ and $\mu_i \sim p(\mu)$, respectively.
\end{assumption}
\begin{assumption}
\label{ass_2}
The  RL algorithm that is to be approximated, as shown in Equations \eqref{eq:rl_map} and \eqref{eq:rl_map_reward}, is converged to the near-optimal value function and policy.
\end{assumption}
Assumption \ref{ass_1} is a common assumption on the task distribution in meta-learning methods \cite{finn2017model}. While Assumption \ref{ass_2} appears strong, it is related to the common assumption made in imitation learning where the learner has access to expert demonstrations  \cite{ross2011reduction, ho2016generative}. Nevertheless, we empirically show that Assumption \ref{ass_2} can be relaxed to an extent in practice, while still achieving strong zero-shot performance, as shown in Section~\ref{sec:results}.

\begin{figure}[t!]
    \centering
    \includegraphics[width=0.49\textwidth]{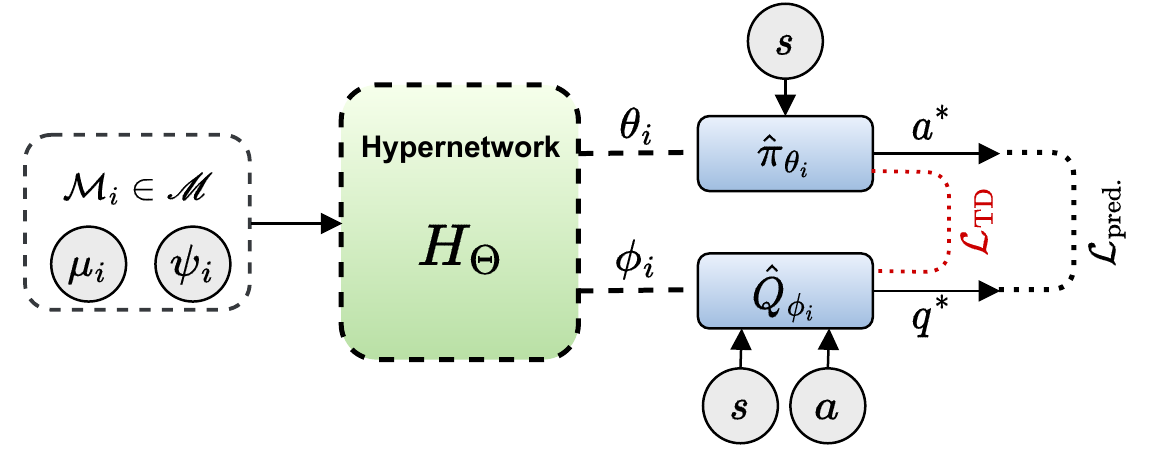}
    \caption{Diagram of our learning framework for universal approximation of RL solutions. Given reward parameters $\psi_i$ and dynamics parameters $\mu_i$, the hypernetwork $H_\Theta$ generates weights of the approximated near-optimal policy $\hat{\pi}_\theta$ and value function $\hat{Q}_\phi$. The only learnable parameters are $\Theta$.}
    \label{fig:hypenet}
\end{figure}

\begin{algorithm}[b!]
    \caption{HyperZero}
    \label{alg:approx_rl}
    \textbf{Inputs}: Parameterized reward function $R_\psi$ and transition dynamics $\T_\mu$, distribution $p(\psi)$ and $p(\mu)$ over parameters, hypernetwork $H_\Theta$, main networks $\hat{\pi}_\theta$ and $\hat{Q}_\phi$. \\
    \textbf{Hyperparameters:} RL algorithm, learning rate $\alpha$ of hypernetwork $H_\Theta$, number of tasks $N$.
\begin{algorithmic}[1]
    \State Initialize dataset $\D$ of near-optimal trajectories 
    \For{$i = 1$ to $N$}
        \State Sample MDP $\M_i \in \mathscr{M}$: $\psi_i \sim p(\psi), \mu_i \sim p(\mu_i)$
        \State Obtain $\pi_i^*$ and $Q^*_i$ of $\M_i \in \mathscr{M}$ with an RL solver
        \State Store near-optimal trajectories $\tau_i^*$: $\D \leftarrow \D \cup \{\tau^*_i \} $
    \EndFor
    \While{not done}
        \State Sample mini-batch $\langle \! \psi_i, \mu_i, s, a^*, s', r, q^*\! \rangle \sim \D$
        \State Generate $\hat{\pi}_{\theta_i}$ and $\hat{Q}_{\phi_i}$: $[\theta_i; \phi_i] = H_\Theta(\psi_i, \mu_i)$
        \State $\Theta \leftarrow \argmin \mathcal{L}_\text{pred.}(\Theta) + \mathcal{L}_\text{TD}(\Theta)$ \Comment{Eqn. (\ref{eq:hypenet_loss}-\ref{eq:td_loss})}
    \EndWhile
\end{algorithmic}
\end{algorithm}

\begin{figure*}[b!]
    \centering
    \begin{subfigure}[b]{0.33\textwidth}
        \includegraphics[width=\textwidth]{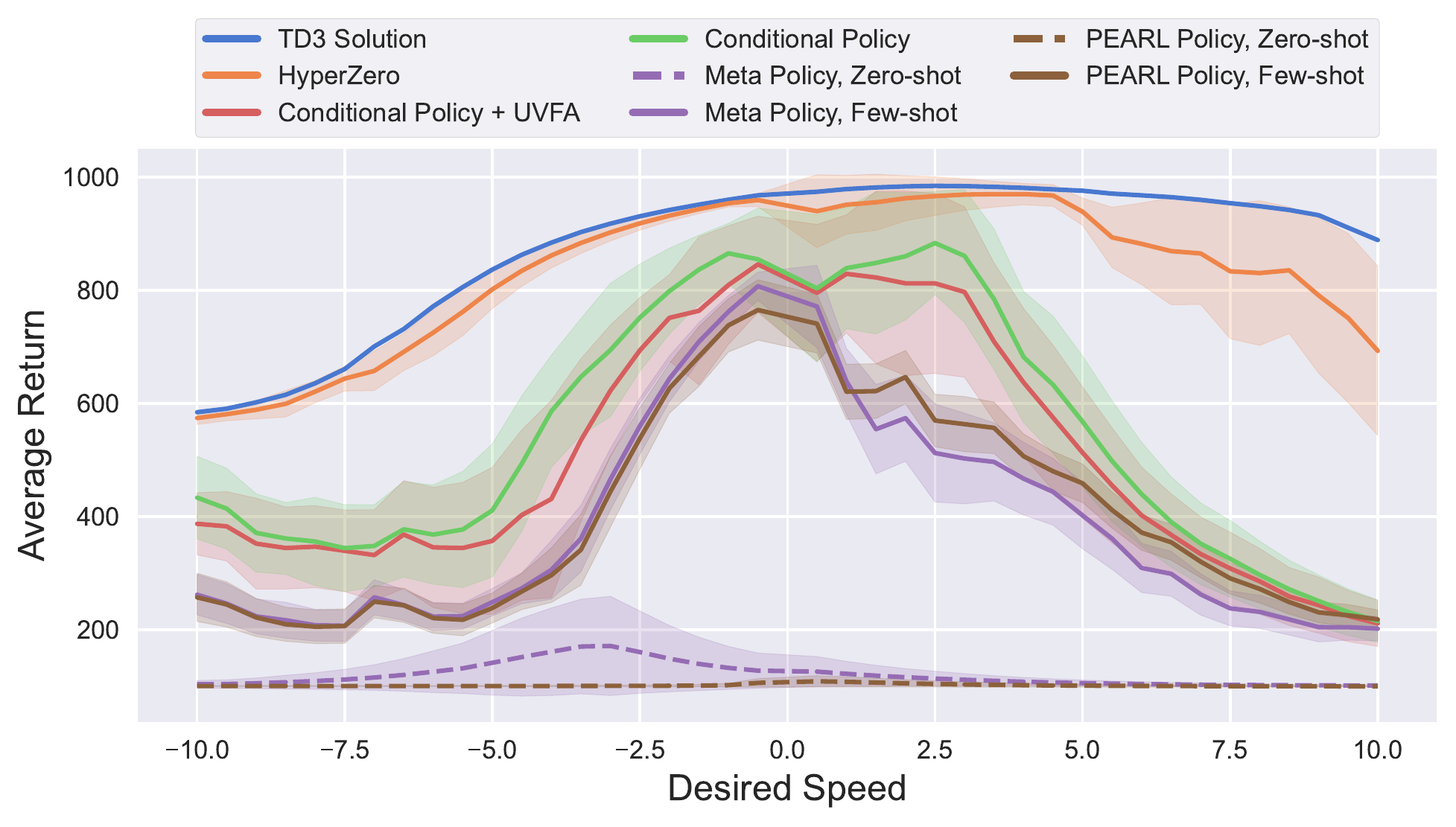}
        \caption{Cheetah environment.}
    \end{subfigure}
    \hfill
    \begin{subfigure}[b]{0.33\textwidth}
        \includegraphics[width=\textwidth]{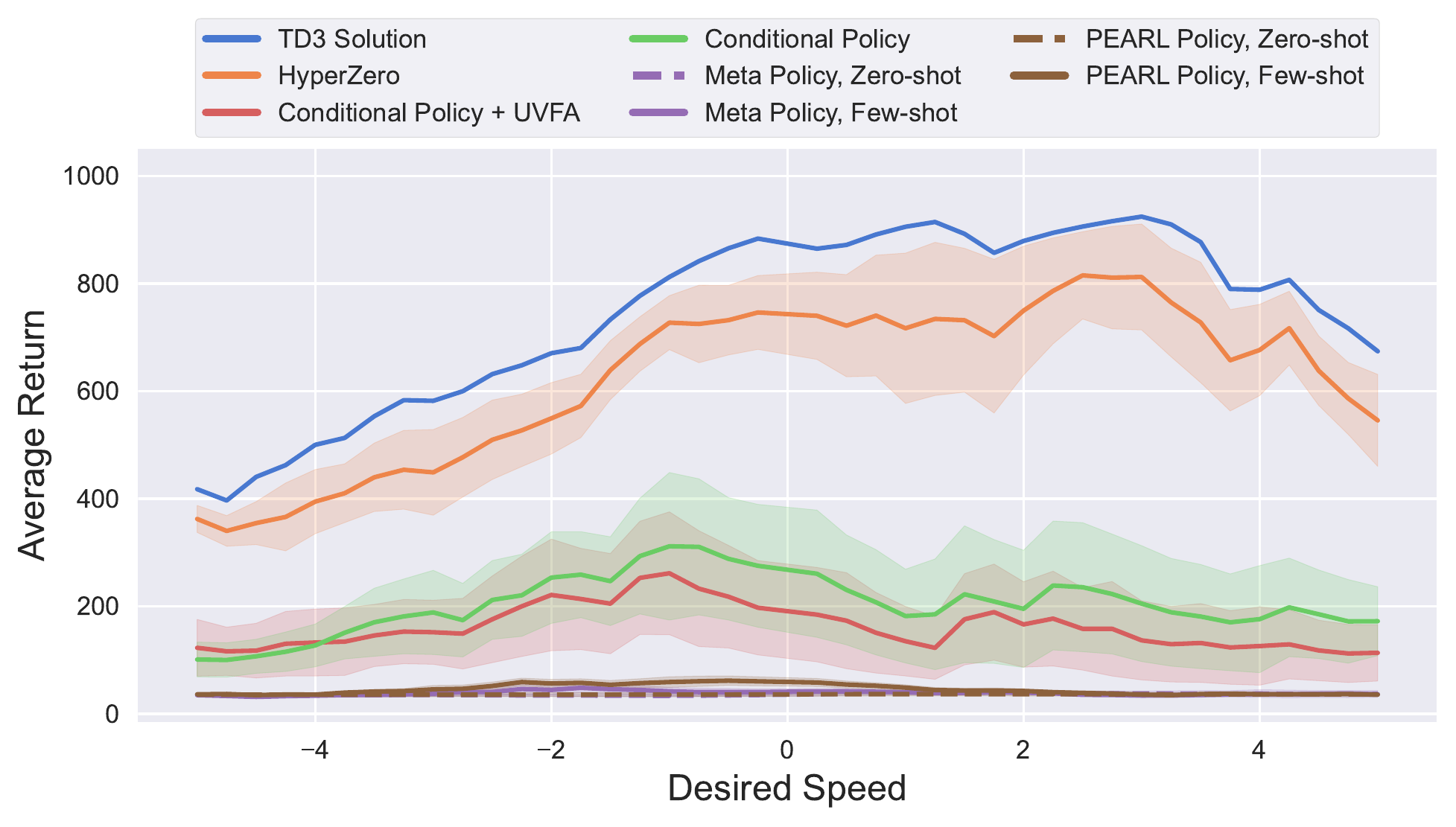}
        \caption{Walker environment.}
    \end{subfigure}
    \hfill
    \begin{subfigure}[b]{0.33\textwidth}
        \includegraphics[width=\textwidth]{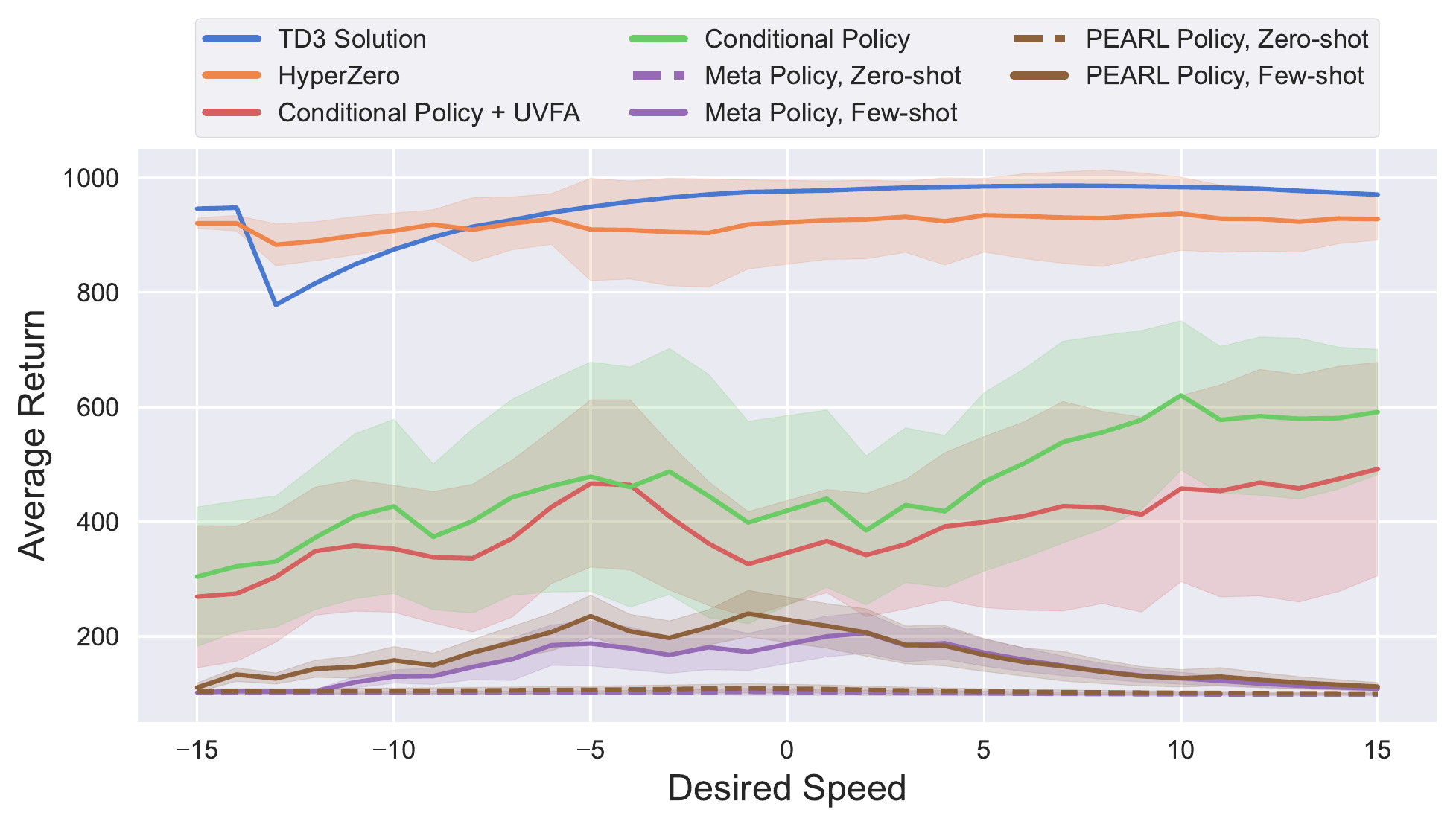}
        \caption{Finger environment.}
    \end{subfigure}
    \caption{Zero-shot transfer to new \textbf{reward} settings on DM control environments, obtained on 5 seeds for random split of train/test tasks. Solid lines present the mean and shaded regions present the standard deviation of the average return across the seeds. Horizontal axis shows the desired speed, which is a function of the reward parameters $\psi_i$.}
    \label{fig:approx_rew}
\end{figure*}

\begin{figure*}[t!]
    \centering
    \begin{subfigure}[b]{0.33\textwidth}
        \includegraphics[width=\textwidth]{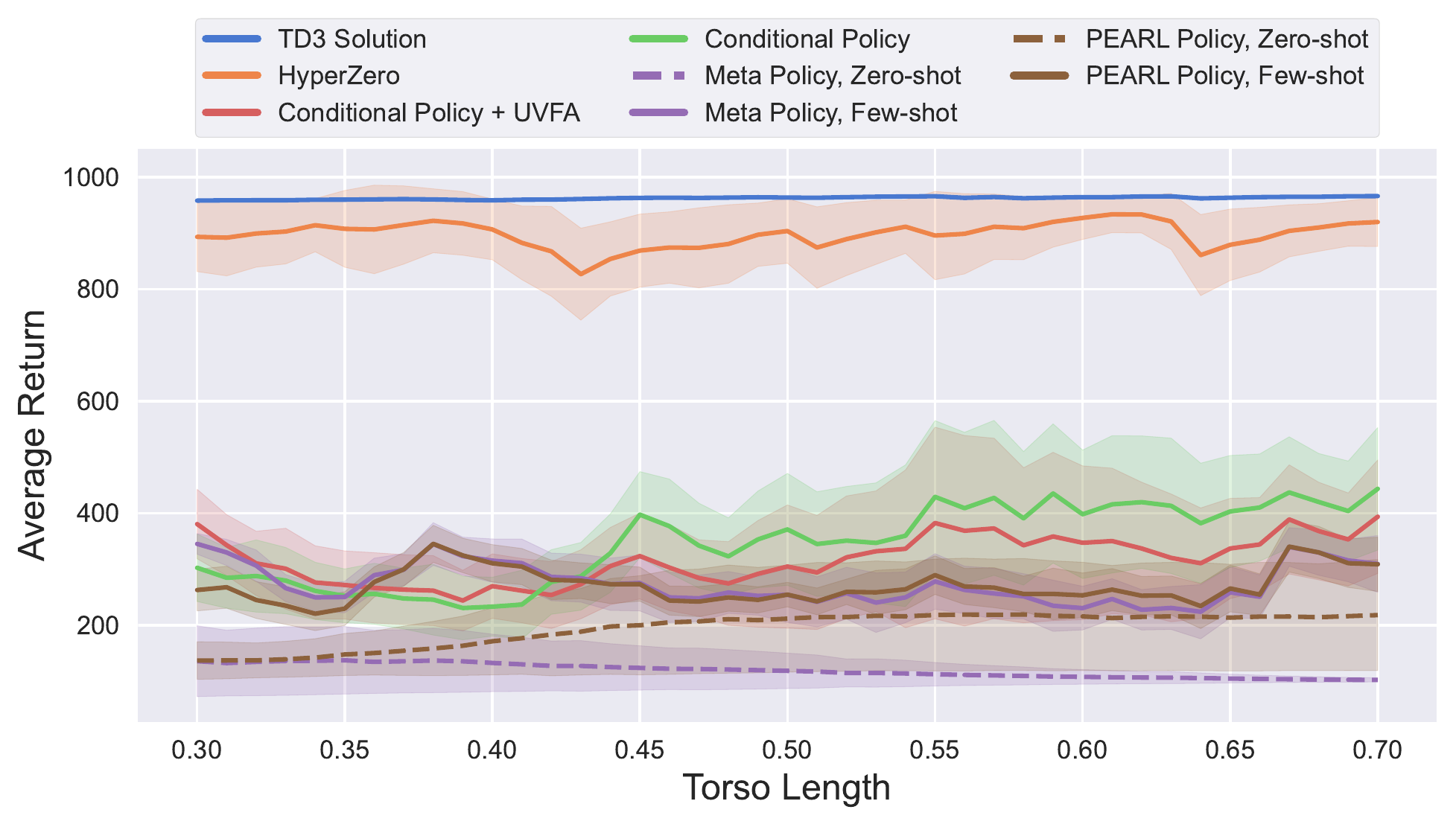}
        \caption{Cheetah environment.}
    \end{subfigure}
    \hfill
    \begin{subfigure}[b]{0.33\textwidth}
        \includegraphics[width=\textwidth]{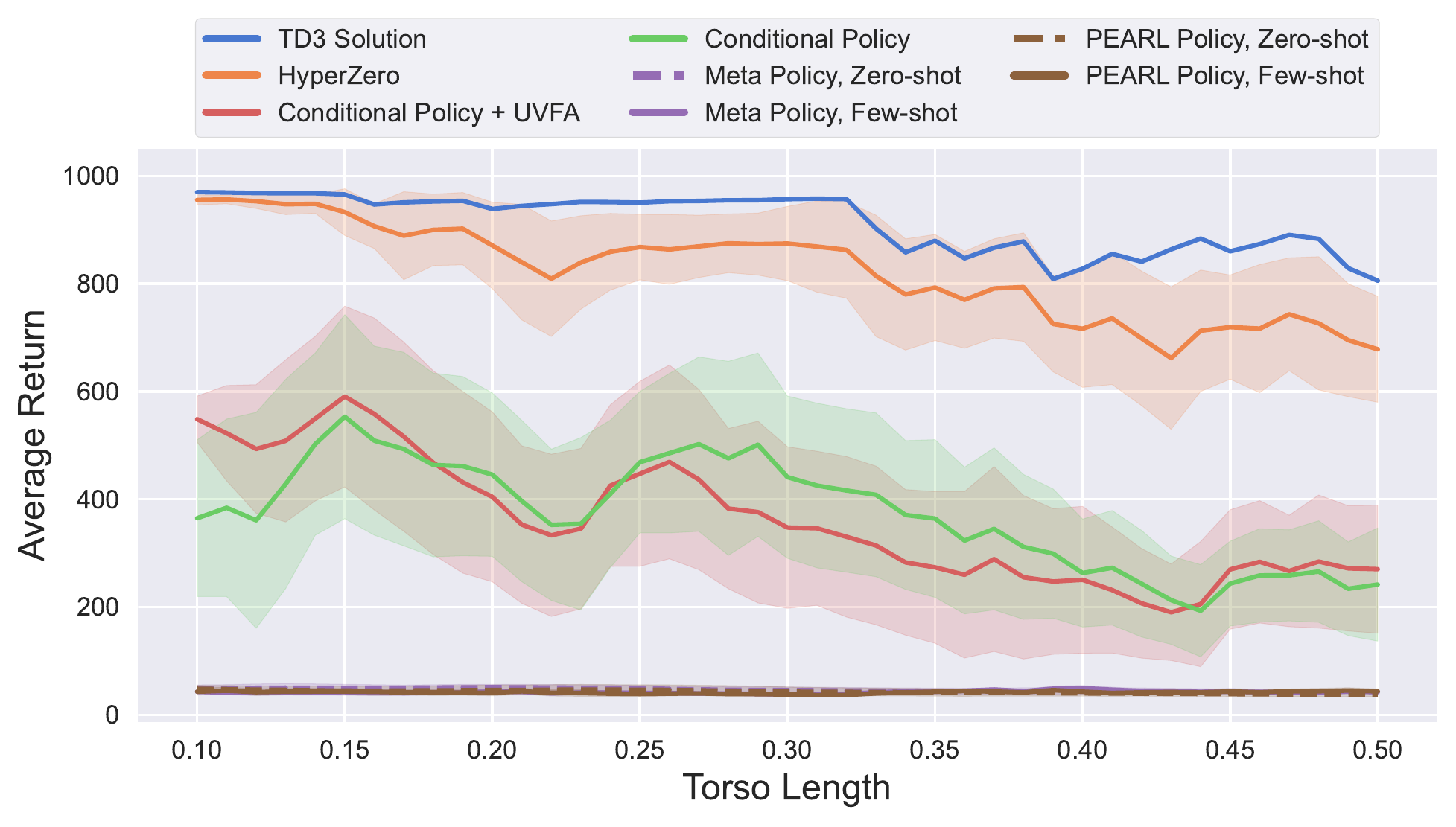}
        \caption{Walker environment.}
    \end{subfigure}
    \hfill
    \begin{subfigure}[b]{0.33\textwidth}
        \includegraphics[width=\textwidth]{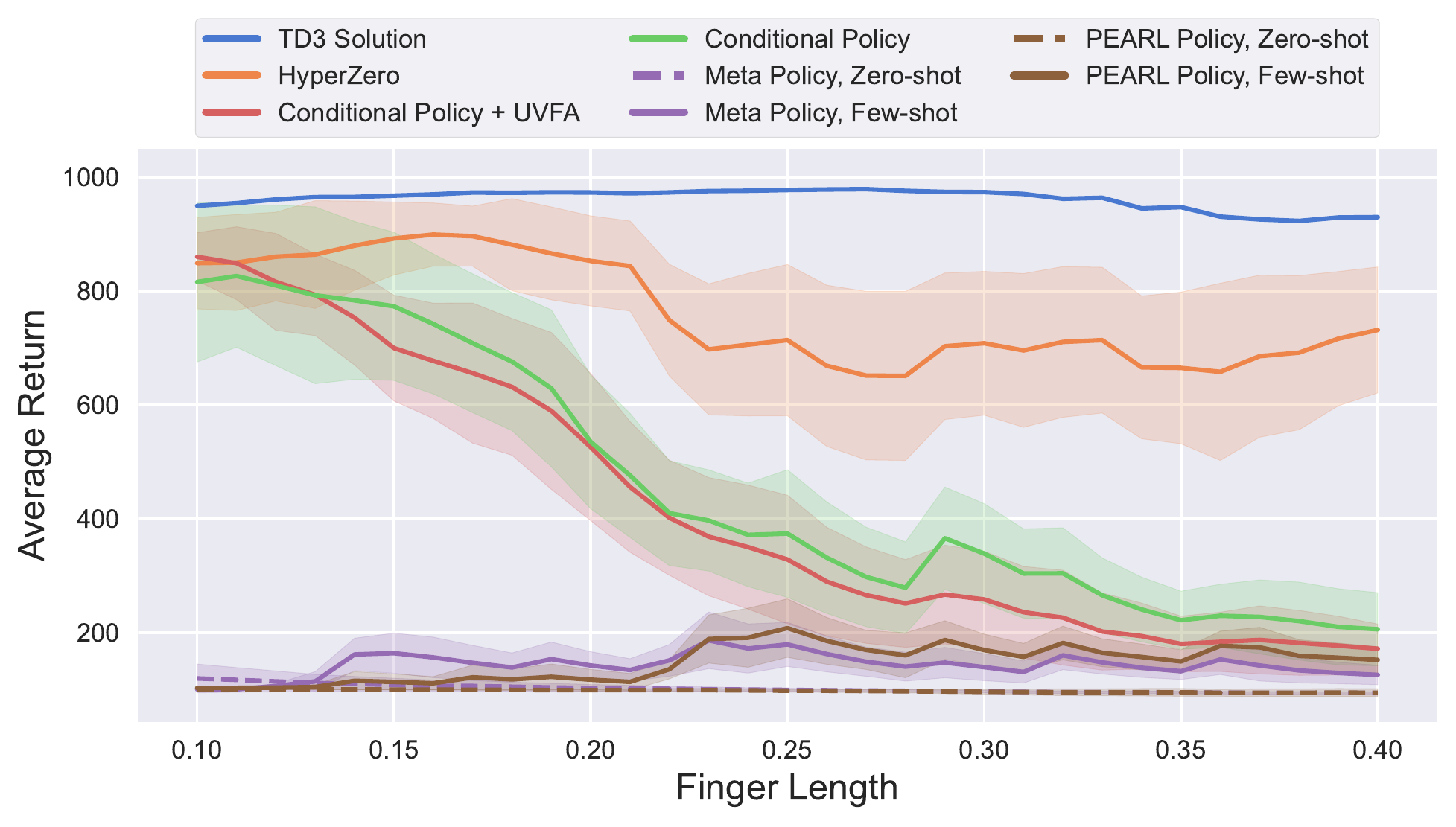}
        \caption{Finger environment.}
    \end{subfigure}
    \caption{Zero-shot transfer to new \textbf{dynamics} settings on DM control environments, obtained on 5 seeds for random split of train/test tasks. Solid lines present the mean and shaded regions present the standard deviation of the average return across the seeds. Horizontal axis shows the value of dynamics parameter $\mu_i$; that is torso length for the cheetah and walker, and finger length for the finger. Notably, a change in the shape of the geometry results in changes in the weight and inertia parameters.}
    \label{fig:approx_dyn}
\end{figure*}

Importantly, we assume no prior knowledge on the structure of the RL algorithm nor on the nature (stochastic or deterministic) of the policy that were used to generate the data. Notably, since the optimal policy of a given MDP is deterministic \cite{puterman2014markov}, we can parameterize the approximated near-optimal policy $\hat{\pi}_\theta$ as a deterministic function $\hat{\pi}_\theta: \s \to \A$, without any loss of optimality.

We propose to use hypernetworks \cite{ha2016hypernetworks} for solving this approximation problem. Conditioned on the parameters $\psi_i, \mu_i$ of an MDP $\M_i \!\in\! \mathscr{M}$, as shown in Figure \ref{fig:hypenet}, the hypernetwork $H_\Theta$ generates weights of the approximated near-optimal policy $\hat{\pi}_\theta$ and action-value function $\hat{Q}_\phi$. Following the literature on hypernetworks, we refer to the generated policy and value networks as \emph{main} networks.

Consequently, the hypernetwork is trained via minimizing the error for predicting the near-optimal action and values by forward passing the main networks:
\begin{align}
    \label{eq:hypenet_loss}
    \mathcal{L}_\text{pred.}(\Theta) = &\E_{(\psi_i, \mu_i, s, a^*, q^*) \sim \D} \Big[(\hat{Q}_{\phi_i}(s, a^*) - q^*)^2 \Big] \nonumber \\
    &+ \E_{(\psi_i, \mu_i, s, a^*) \sim \D} \Big[(\hat{\pi}_{\theta_i}(s) - a^*)^2 \Big]
\end{align}
where $[\theta_i; \phi_i] = H_\Theta(\psi_i, \mu_i)$ and $\D$ is the dataset of near-optimal trajectories collected from the family of MDPs $\mathscr{M}$. Notably, this training paradigm effectively decouples the problem of learning optimal values/actions from the problem of learning the mapping of MDP parameters to the space of optimal value functions and policies. Thus, as observed in other works on hypernetworks \cite{ha2016hypernetworks, galanti2020modularity, von2019continual, faccio2022goal}, this level of modularity results in a simplified and more efficient learning.

\subsection{Temporal Difference Regularization}
A key challenge in using supervised learning approaches for function approximation in deep RL is the temporal correlation existing within the samples, which results in the violation of the i.i.d. assumption. Common practices in deep RL for stabilizing the learning is to use a target network to estimate the temporal difference (TD) error \cite{lillicrap2015continuous,mnih2013playing}. In this paper, we propose a  novel regularization technique based on the TD loss to stabilize the training of the hypernetwork for zero-shot transfer learning. 

As stated in Assumption \ref{ass_2}, we assume having access to near-optimal RL solutions that were generated from a converged RL algorithm. As a result, our framework differs from the works on imitation learning \cite{ross2011reduction, bagnell2015invitation, ho2016generative} since samples satisfy the Optimal Bellman equation of the underlying MDP $\M_i \!\in\! \mathscr{M}$ and, more importantly, we have access to the near-optimal action-values $q^*$ for a given transition sample $\langle s, a^*, s', r\rangle$.

Therefore, we propose to use the TD loss to regularize the approximated critic $\hat{Q}_\phi$ by moving the predicted target value towards the current value estimate, which is obtainable from the ground-truth RL algorithm:
\begin{align}
    \loss_\text{TD}(\Theta) = &\E_{(\psi_i, \mu_i, s, a^*, s', r, q^*) \sim \D} \Big[\!(r \!+\! \gamma \hat{Q}_{\phi_i}(s', \overline{a'}) -  q^* )^2 \Big]
    \label{eq:td_loss}
\end{align}
where $\overline{a'}$ is obtained from the approximated deterministic policy $\hat{\pi}_\theta(s')$ with stopped gradients. Note that our application of the TD loss differs from that of standard function approximation in deep RL \cite{mnih2013playing, lillicrap2015continuous}; instead of moving the current value estimate towards the target estimates, our TD loss moves the target estimates towards the current estimates. While this relies on Assumption \ref{ass_2}, we show that in practice applying the TD loss is beneficial as it enforces the approximated policy and critic to be consistent with respect to the Bellman equation. 
Algorithm \ref{alg:approx_rl} shows the pseudo-code of our learning framework.

\begin{figure*}[b!]
    \centering
    \begin{subfigure}[b]{0.32\textwidth}
        \includegraphics[width=\textwidth]{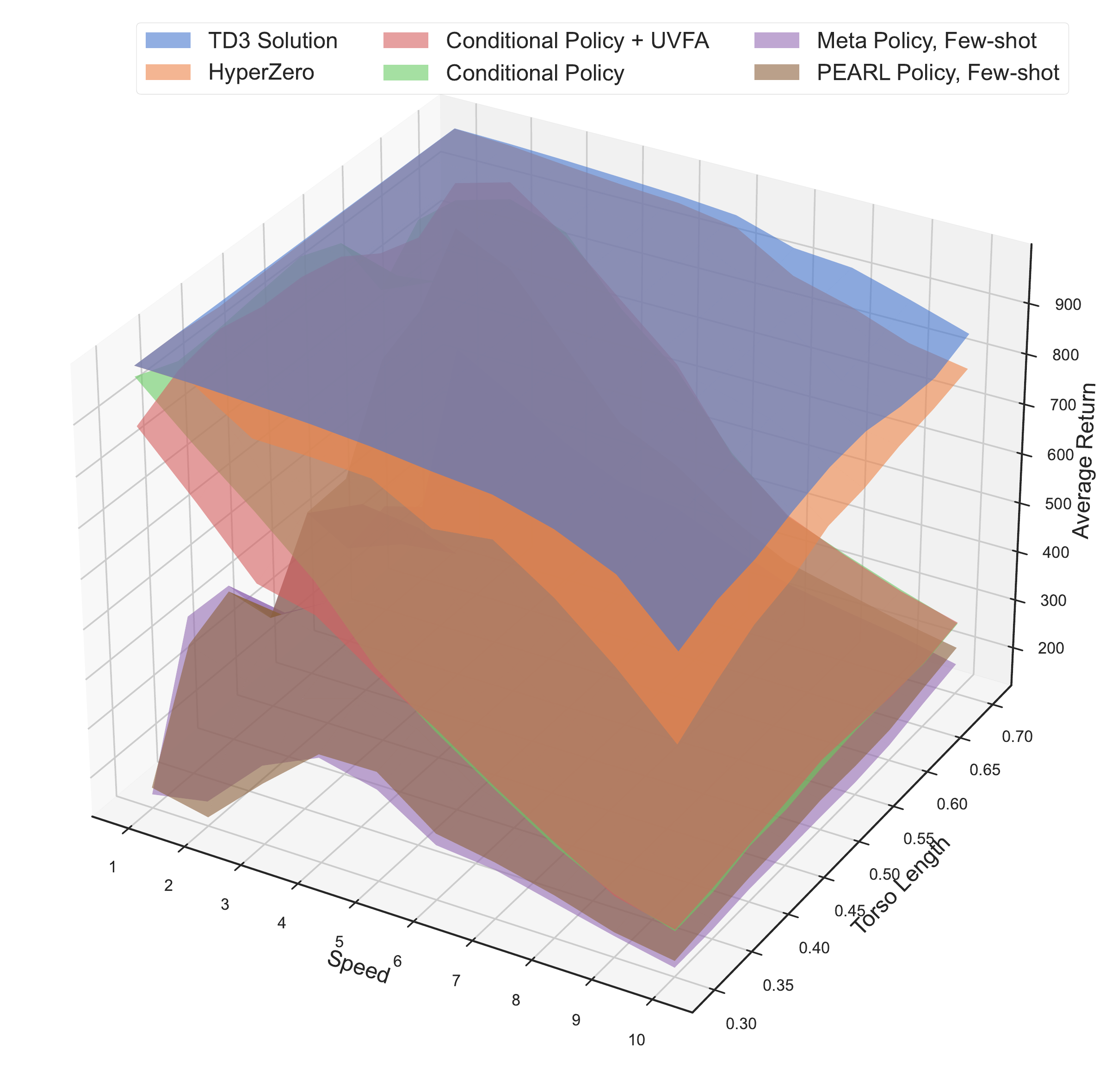}
        \caption{Cheetah environment.}
    \end{subfigure}
    \hfill
    \begin{subfigure}[b]{0.32\textwidth}
        \includegraphics[width=\textwidth]{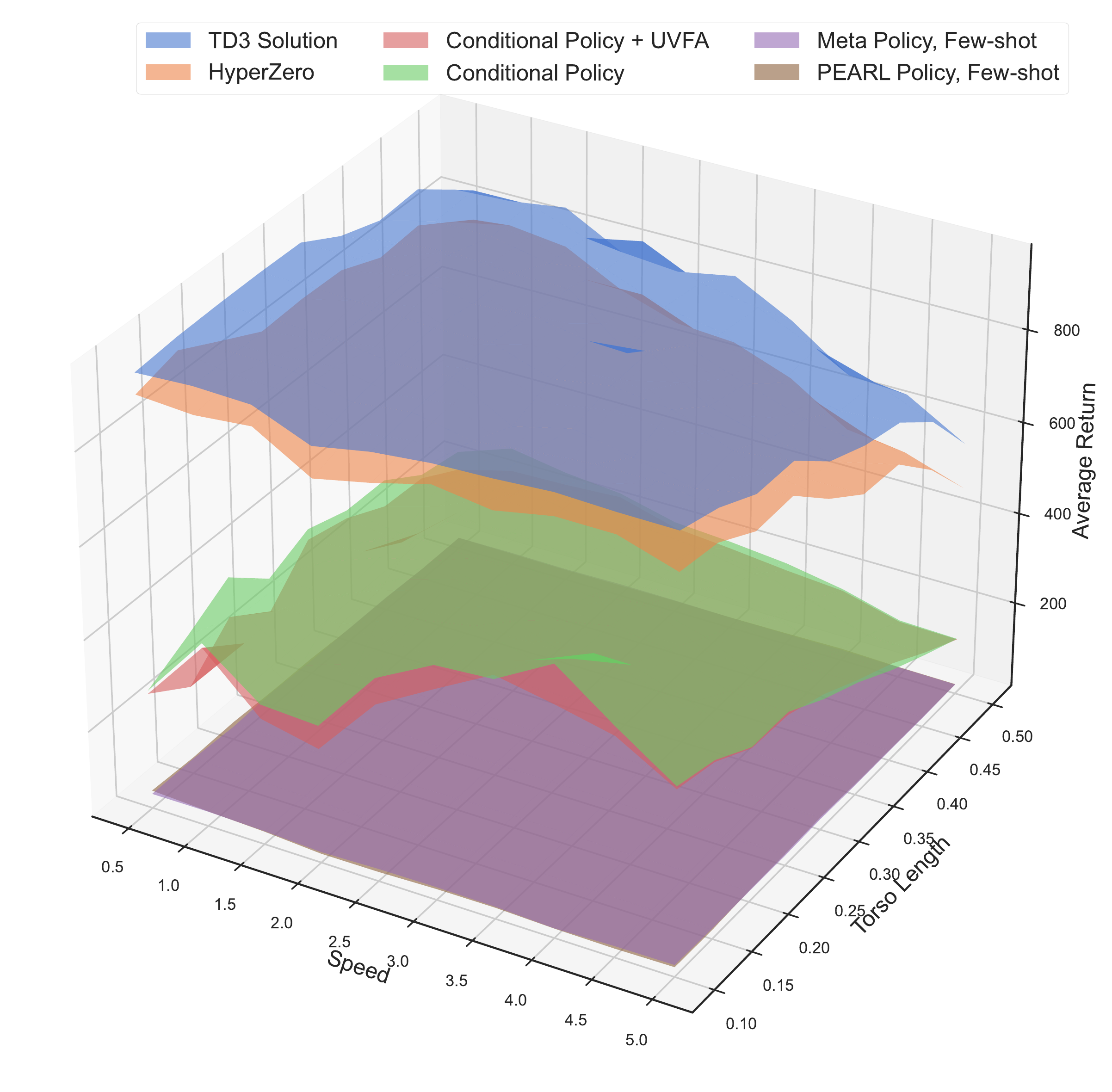}
        \caption{Walker environment.}
    \end{subfigure}
    \hfill
    \begin{subfigure}[b]{0.32\textwidth}
        \includegraphics[width=\textwidth]{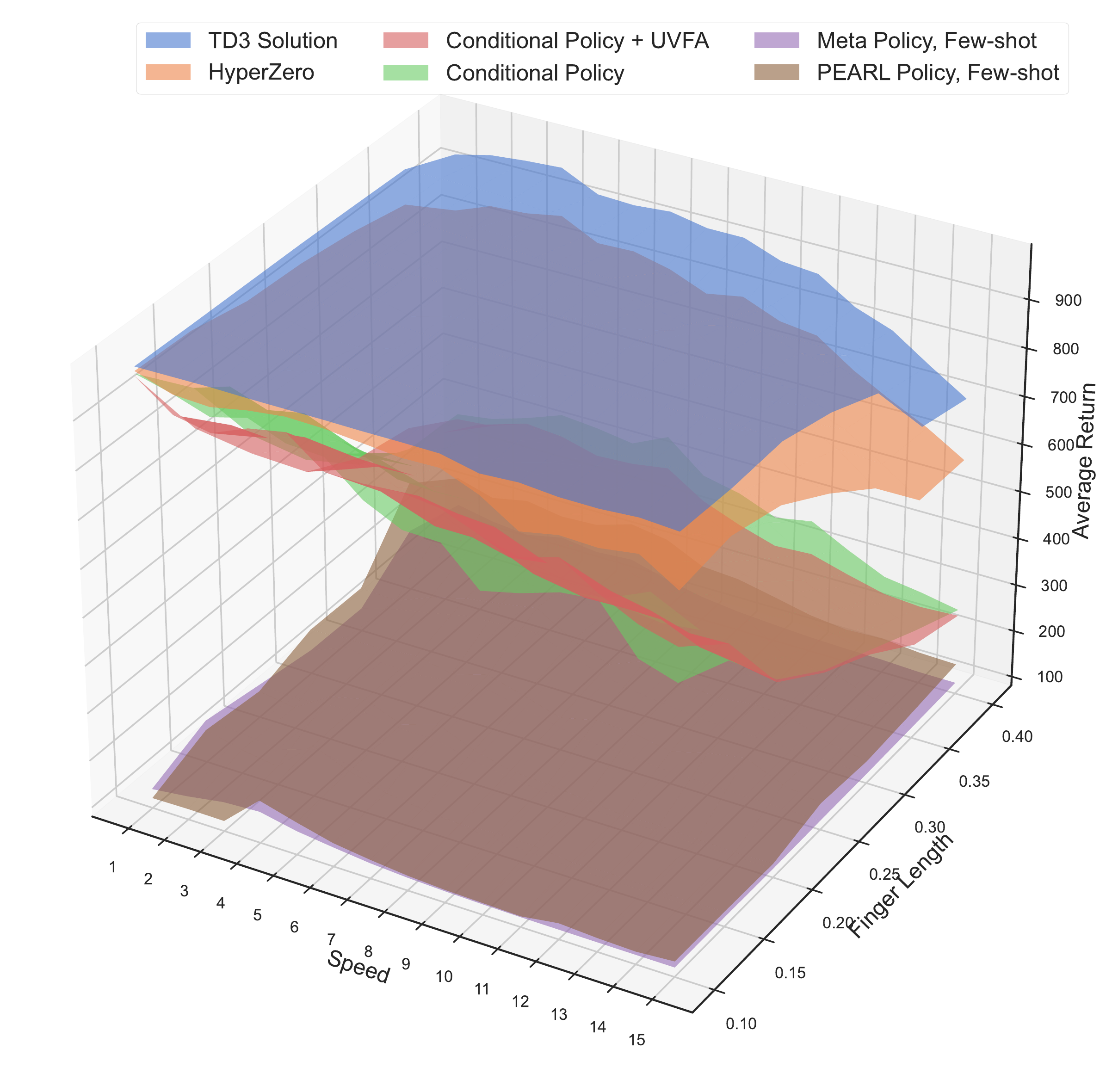}
        \caption{Finger environment.}
    \end{subfigure}
    \caption{Zero-shot transfer to new \textbf{reward and dynamics} settings on DM control environments, obtained on 5 seeds for random split of train/test tasks. Each surface present the mean of the average return across the seeds. X-axis shows the desired speed, which is a function of the reward parameters $\psi_i$, while Y-axis shows the value of the dynamics parameter $\mu_i$. The surfaces are smoothed for visual clarity. 2D plots of these 3D diagrams are presented in Appendix \ref{supp:additional_results} for better comparison.}
    \label{fig:approx_rew_dyn}
\end{figure*}

\begin{figure*}[t!]
    \centering
    \begin{subfigure}[b]{0.49\textwidth}
        \includegraphics[width=\textwidth]{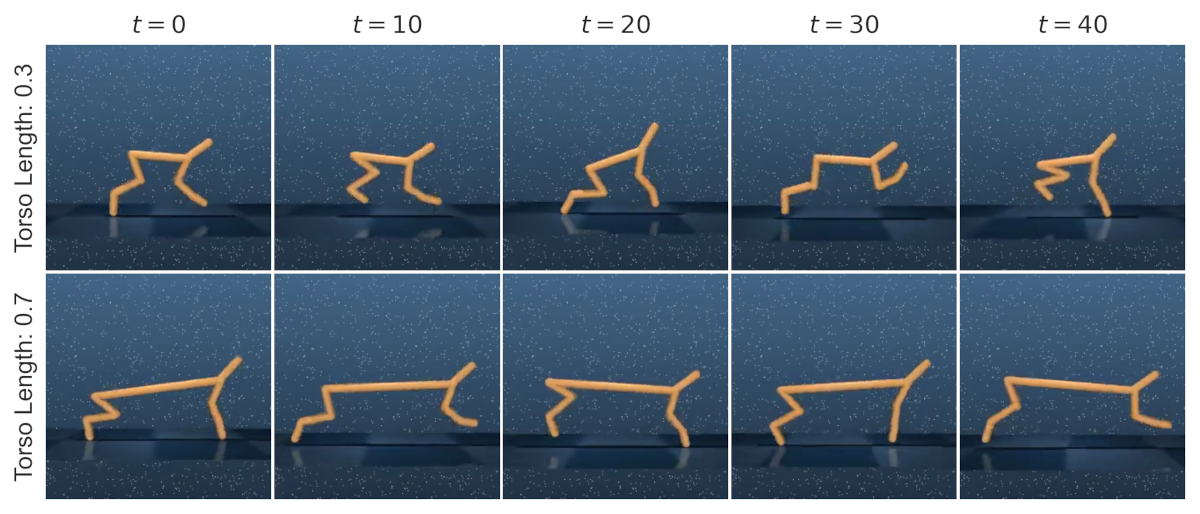}
        \caption{Cheetah environment with different torso lengths.}
    \end{subfigure}
    \hfill
    \begin{subfigure}[b]{0.49\textwidth}
        \includegraphics[width=\textwidth]{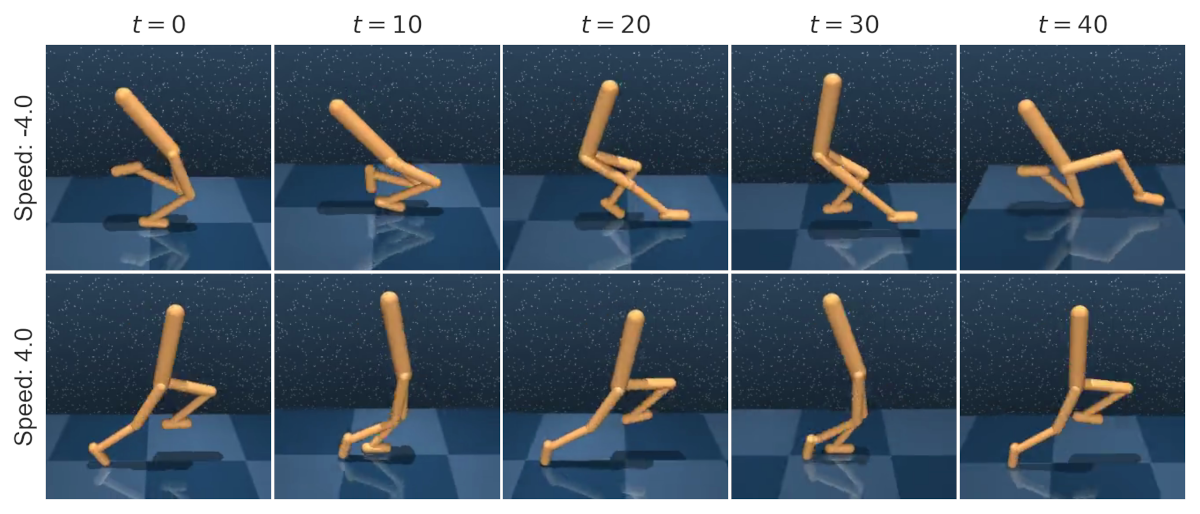}
        \caption{Walker environment with different desired speeds.}
    \end{subfigure}
    \caption{Rollout of a trained HyperZero on different task parameters. \textbf{(a)} The trained HyperZero is used to rollout the cheetah environment with torso lengths of 0.3 and 0.7. \textbf{(c)} The trained HyperZero is used to rollout the walker environment with desired speeds of -4 and +4. Additional results are in Appendix \ref{supp:example_behavior}.}
    \label{fig:example_behaviours_main}
\end{figure*}

\section{Evaluation}
\label{sec:results}
We evaluate our proposed method, referred to as \emph{HyperZero} (\textbf{hyper}networks for \textbf{zero}-shot transfer) on a series of challenging continuous control tasks from DeepMind Control Suite. The primary goal in our experiments is to study the zero-shot transfer ability of the approximated RL solutions to novel dynamics and rewards settings. 

\subsection{Experimental Setup}
\paragraph{Environments.} We use three challenging environments for evaluation: cheetah, walker, and finger. For an easier visualization and realization of reward parameters, in all cases the reward parameters correspond to the desired speed of the motion which consists of both negative (moving backward) and positive (moving forward) values. Depending on the environment, dynamics changes correspond to changes in a body size and its weight/inertia. Full details of the environments and their parameters are in Appendix \ref{supp:exp_details}.

\paragraph{RL Training and Dataset Collection.} We use TD3 \cite{fujimoto2018addressing} as the RL algorithm that is to be approximated. Each MDP $\M_i \!\in\! \mathscr{M}$, generated by sampling $\psi_i \!\sim\! p(\psi)$ and $\mu_i \!\sim\! p(\mu)$, is used to independently train a standard TD3 agent on proprioceptive states for 1 million steps. Consequently, the final solution is used to generate 10 rollouts to be added to the dataset $\mathcal{D}$. Learning curves for the RL solutions are in Appendix \ref{supp:rl_solutions}. As these results show, in some instances, the RL solution is not fully converged after 1 million steps. Despite this, HyperZero is able to approximate the mapping reasonably well, thus indicating Assumption \ref{ass_2} can be relaxed to an extent in practice.

\paragraph{Train/Test Split of the Tasks.} To reliably evaluate the zero-shot transfer abilities of HyperZero to novel reward/dynamics settings against the baselines, and to rule out the possibility of selective choosing of train/test tasks, we randomly divide task settings into train ($\%85$) and test ($\%15$) sets. We consequently report the mean and standard deviation of the average return obtained on 5 seeds.

\paragraph{Baselines.} We compare HyperZero against common baselines for multitask and meta learning:
\begin{enumerate}
    \item Context-conditioned policy; trained to predict actions, similarly to imitation learning methods. 
    \item Context-conditioned policy paired with UVFA \cite{schaul2015universal}; trained to predict actions and values. It further benefits from using our proposed TD loss $\loss_\text{TD}$, similarly to HyperZero.
    \item Context-conditioned meta policy; trained with MAML \cite{finn2017model} to predict actions and evaluated for both zero-shot and few-shot transfer. Our context-conditioned meta-policy can be regarded as an adaptation of PEARL \cite{rakelly2019efficient} in which the inferred task is substituted by the ground-truth task. 
    \item PEARL \cite{rakelly2019efficient} policy; trained to predict actions. Unlike other baselines, PEARL does not assume access to the MDP context and instead it infers the the context from states and actions. 
\end{enumerate}
Notably, since MAML and PEARL are known to perform poorly for zero-shot transfer, we evaluate the meta policy for both zero-shot and few-shot transfers. In the latter, prior to evaluation, the meta policy is finetuned with near-optimal trajectories of the test MDP $\M_i$ generated by the actual RL solution. 

Finally, for a fair comparison with hypernetworks, all methods follow the same two-stage training paradigm described in Section \ref{sec:problem_form}, have a learnable task embedding, and share the same network architecture. Full implementation details are in Appendix \ref{supp:hyperparam_details}.

\subsection{Results}
\subsubsection{Zero-shot Transfer.} 
We compare the zero-shot transfer of HyperZero against the baselines in the three cases of changed rewards, changed dynamics, and simultaneously changed rewards and dynamics; results are shown in Figures \ref{fig:approx_rew}, \ref{fig:approx_dyn}, and \ref{fig:approx_rew_dyn}, respectively. Additional results are in Appendix \ref{supp:additional_results}. As suggested by these results, in all environments and transfer scenarios, HyperZero significantly outperforms the baselines, demonstrating the effectiveness of our learning framework for approximating an RL algorithm as a mapping from a parameterized MDP $\M_i$ to a near-optimal policy $\pi^*_i$ and action-value function $Q^*_i$.

Importantly, the context-conditioned policy (paired with UVFA) consists of all the major components of HyperZero, including near-optimal action and value prediction, and TD regularization. As a result, the only difference is that HyperZero learns to generate policies conditioned on the context which is in turn used to predict actions, while the context-conditioned policy learns to predict actions conditioned on the context. We hypothesize two main reasons for the significant improvements gained from such use of hypernetworks in our setting. First, aligned with similar observations in the literature \cite{galanti2020modularity, von2019continual}, hypernetworks allow for effective abstraction of the learning problem into two levels of policy (or equivalently value function) generation and action (or equivalently value) prediction.

\begin{figure}[b!]
    \centering
    \begin{subfigure}[b]{0.45\textwidth}
        \includegraphics[width=\textwidth]{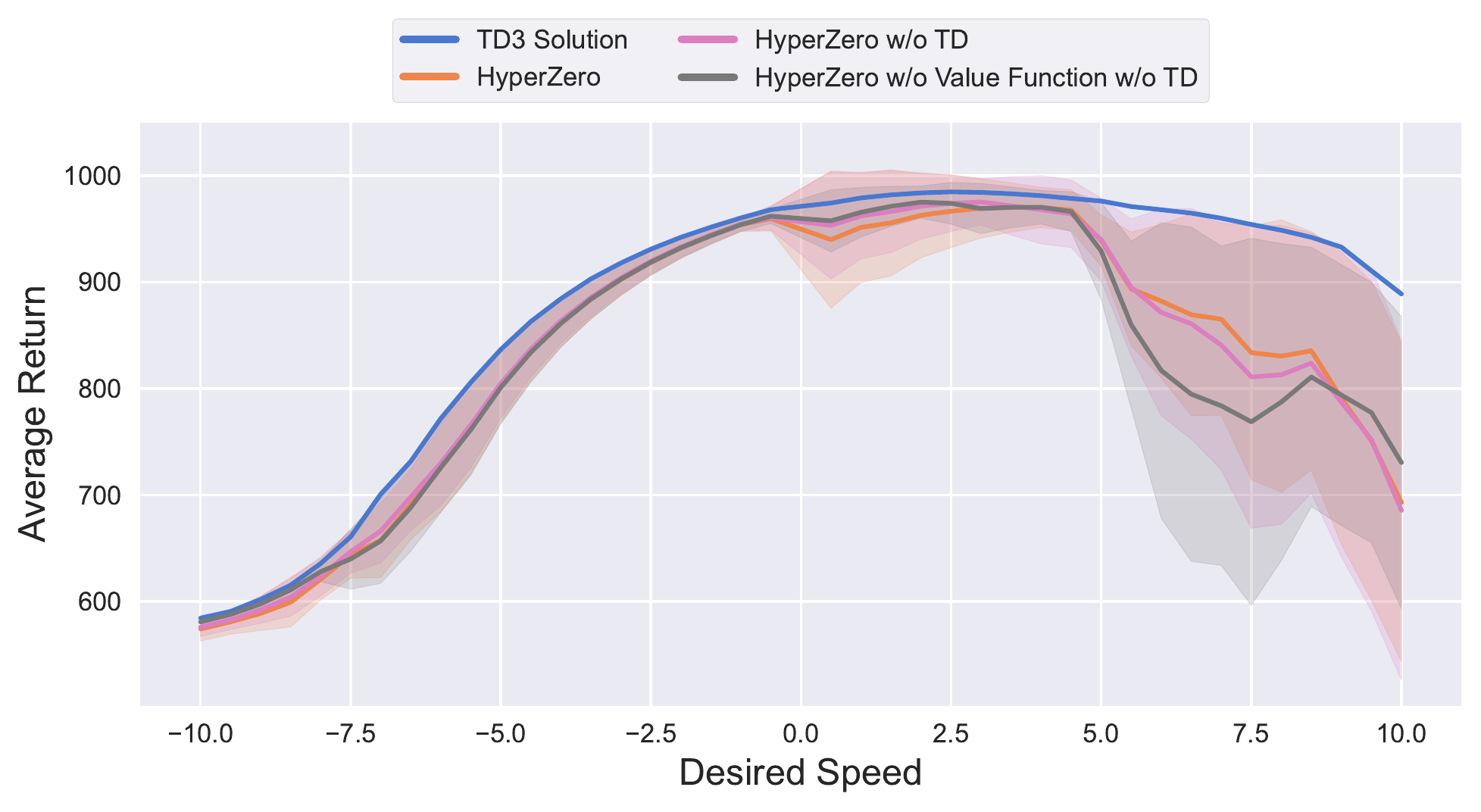}
        \caption{Cheetah environment.}
    \end{subfigure}
    
    \begin{subfigure}[b]{0.45\textwidth}
        \includegraphics[width=\textwidth]{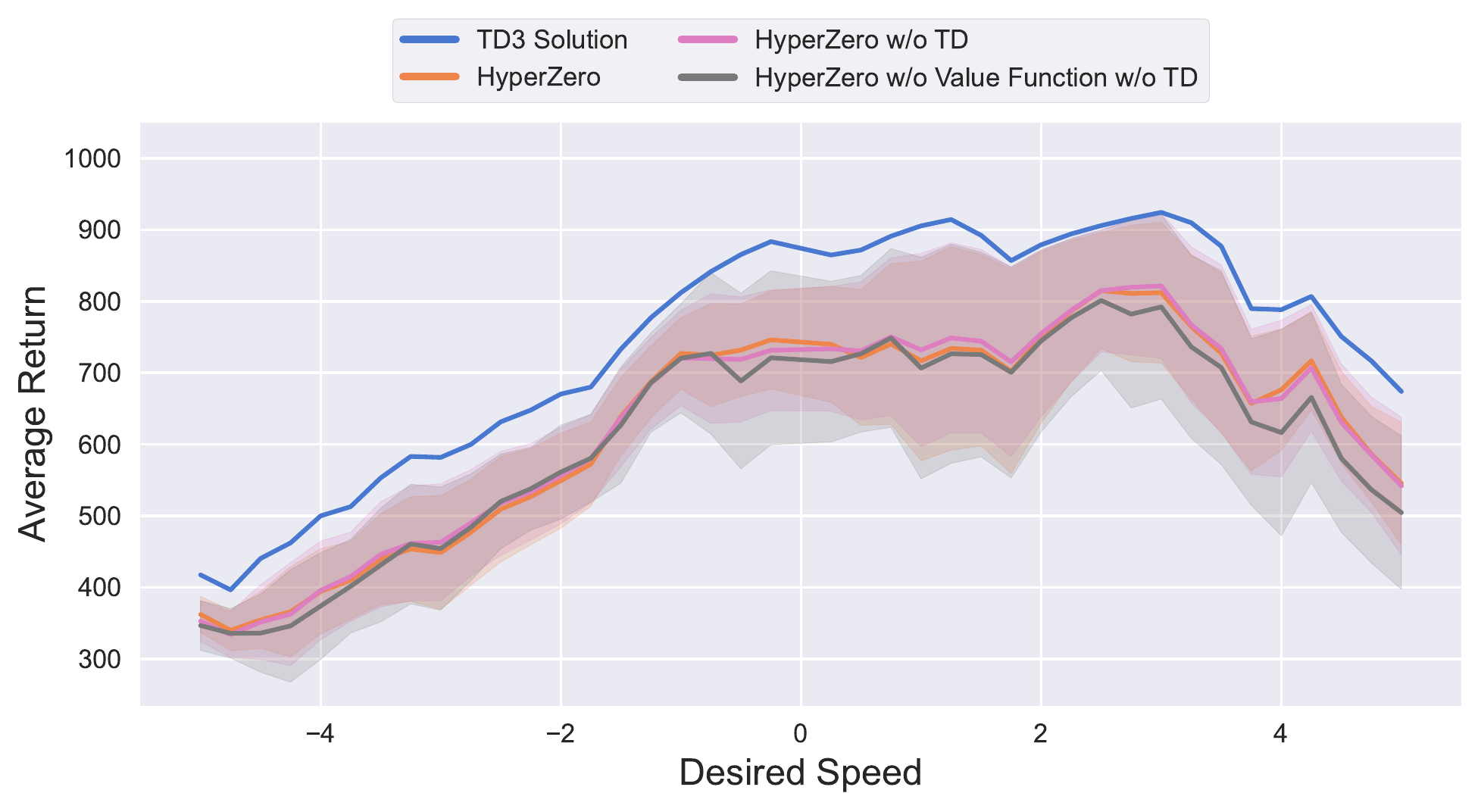}
        \caption{Walker environment.}
    \end{subfigure}
    \caption{Ablation study on the improvements gained from generating the optimal value function and using the TD loss. Results are obtained on 5 seeds for random split of train/test tasks. Solid lines present the mean and shaded regions present the standard deviation of the average return across the seeds. Horizontal axis shows the desired speed, which is a function of the reward parameters $\psi_i$.}
    \label{fig:ablation}
\end{figure}

Second, as  hypernetworks are used to learn the mapping from MDP parameters to the space of policies, that is $(\psi_i, \mu_i) \to \pi^*_i$, they achieve generalization across the space of policies. On the contrary, since the context-conditioned policy simultaneously learns the mapping of states and MDP parameters to actions, that is $(s, \psi_i, \mu_i) \to a^*$, it is only able to achieve generalization over the space of actions, as opposed to the more general space of policies.

Finally, due to the strong zero-shot transfer ability of the approximated solution to new rewards and dynamics, one can use it to visualize the near-optimal trajectory $\tau_i^*$ for novel tasks without necessarily training the RL algorithm. A possible application of this approach would be for task visualization or environment design, as well as manual reward shaping. As an example, Figure \ref{fig:example_behaviours_main} shows sample trajectories generated by rolling out trained HyperZero models conditioned on different reward/dynamics parameters. Additional trajectories are in Appendix \ref{supp:example_behavior}.

\subsubsection{Ablation Study on HyperZero Variants.} In Figure \ref{fig:ablation}, we carry out an ablation study on the improvements gained from generating the near-optimal value function and using our proposed TD loss from Equation \eqref{eq:td_loss}. We draw two conclusions from this study; first, generating the action-value function $Q^*_i$ alongside the policy $\pi^*_i$ provides additional learning signal for training the hypernetwork. Furthermore, incorporating the TD loss between the generated policy and action-value function ensures the two generated networks are consistent with one another with respect to the Bellman equation and results in overall better performance and generalization.

While the improvements may appear to be small, we suspect that gains would be larger in visual control problems, as generating the value function will provide a rich learning signal for representation learning. More importantly, the generated value function can have other applications, such as beings used in policy gradient methods for further training the generated policy with environment interactions (offline-to-online RL) \cite{lee2022offline}. While this is left for future work, we wanted to ensure that our framework is capable of generating the value function alongside the policy.

\section{Related Work}
The robustness and generalization of behaviors has long been studied in control and RL. 

\subsubsection{Transfer, Contextual and Meta RL.}
Past work has studied numerous forms of Transfer Learning \cite{JMLR:v10:taylor09a}, where MDP components including the state space, action space, dynamics or reward are modified between the training conducted on one or many source tasks, prior to performance on one or more targets. Depending on the learner's view of sources and targets, the problem is called contextual policy search~\cite{contextualPolicySearch} life-long learning~\cite{abelICML2018}, curriculum learning~\cite{portelasCORL2020}, or meta learning~\cite{finn2017model}, but our particular variant, with an always-observable parameter vector and no chance to train or fine-tune on the target is most aptly named zero-shot contextual RL. Within that problem, a common concern has been how to interpolate in the space of contexts (equivalent to our parameters), while preserving details of the policy-space solution \cite{victorICRA2018}. This is precisely where the power of our hypernetwork architecture extends prior art. 

\subsubsection{Hypernetworks in RL.}
While hypernetworks \cite{ha2016hypernetworks} have been used extensively in supervised learning problems \cite{von2019continual, galanti2020modularity, krueger2017bayesian, zhao2020meta}, their application to RL algorithms remains relatively limited.
Recent work of \citet{sarafian2021recomposing} use hypernetworks to improve gradient estimation of $Q$ functions and policy networks in policy gradient algorithms. In multi-agent RL, hypernetworks are used to generate policies or value functions based on agent properties \cite{rashid2018qmix, iqbal2020ai, iqbal2021randomized, de2020deep, zhou2020learning}. Furthermore, hypernetworks have been used to model an evolving dynamical system in continual model-based RL \cite{huang2021continual}. Related to our approach, \citet{faccio2022goal} use hypernetworks to learn goal-conditioned optimal policies; the key distinguishing factor of our approach is that we focus on zero-shot transfer across a family of MDPs with different reward and dynamics functions, while the method of \citet{faccio2022goal} aims to solve a single goal-conditioned MDP.


\paragraph{Upside Down RL.}
Upside down RL (UDRL) is a redefinition of the RL problem transforming it into a form of supervised learning. UDRL, rather than learning optimal policies using rewards, teaches agents to follow commands. This method maps input observations as commands to action probabilities with supervised learning conditioned on past experiences \cite{srivastava2019training, schmidhuber2019reinforcement}. Related to this idea are offline RL models that use sequence modeling as opposed to supervised learning to model behavior \cite{janner2021offline, chen2021decision}. Similarly to UDRL, many RL algorithms incorporate the use of supervised learning in their model \cite{schmidhuber2015deep,rosenstein2004supervised}. One such technique is hindsight RL in which commands correspond to goal conditions \cite{andrychowicz2017hindsight, rauber2017hindsight, harutyunyan2019hindsight}. Another approach is to use forward models as opposed to the backward ones used in UDRL \cite{arjona2019rudder}. Recently, \citet{faccio2022goal} propose a method that evaluates generated polices in the command space rather than optimizing a single policy for achieving a desired reward.

\section{Conclusion}
This paper has described an approach, named HyperZero, which learns to generalize optimal behavior across a family of tasks. By training on the full RL solutions of training tasks, including their optimal policy and value function parameters, the hypernetworks used in our architecture are trained to directly output the parameters of complex neural network policies capable of solving unseen target tasks. This work extends the performance of zero-shot generalization over prior approaches. Our experiments demonstrate that our zero-shot behaviors achieve nearly full performance, as defined by the performance of the optimal policy recovered by an RL learner training for a large amount of iterations on the target task itself. 

Due to the strong generalization of our method, with minimal test-time computational requirements, our approach is suitable for deployment in live systems. We also highlight the opportunity for human-interfaces and exploration of RL solutions. In short, this new level of rapid, but powerful, general behavior can provide significant opportunity for the practical deployment of RL-learned behavior in the future.

\bibliography{refs.bib}

\newpage
\appendix
\onecolumn

\section{Details of the Environments and Data Collection}
\label{supp:exp_details}
In this section, we provide the full details of the custom environments and the data collection procedure to supplement the results of Section 4.1 

\subsection{Environment Details}
\label{supp:env_details}
We use three challenging environments for evaluation: cheetah, walker, finger. All environments are derived from DeepMind Control Suite by enabling a modular approach for modifying the reward and dynamics parameters. \emph{The code for our modular and customizable environments will be publicly available by the time of publication.}

\subsubsection{Reward Parameters.} In all environments, the reward parameter $\psi$ correspond to the the desired speed of the agent in order to allow for easier visualization of the learned behaviour. Table \ref{tab:reward} presents the details of the reward parameters used in the experiments for transfer to novel rewards. Notably, desired speed of 0 is not included due to its trivial solution. 
\begin{table}[h!]
\centering
\begin{tabular}{c|c|c|c|c}
\textbf{Environment} & \textbf{Reward Parameter} $\psi$ & \textbf{Range and Increments}       & \textbf{No. of Samples} $p(\psi)$       &  \textbf{Default of DMC}\\
\hline
\hline
\textbf{Cheetah}     & Desired moving speed    & $[-10, +10]$, 0.5 increments & 40 samples & +10\\
\textbf{Walker}      & Desired moving speed    & $[-5, +5]$, 0.25 increments   & 40 samples  & +1 (walk), +8 (run)\\
\textbf{Finger}      & Desired spinning speed  & $[-15, +15]$, 1.0 increments & 30 samples  & +15
\end{tabular}
\caption{Details of the reward parameters used in experiments for transfer to novel \textbf{reward} settings.}
\label{tab:reward}
\end{table}

\subsubsection{Dynamic Parameters.} In all environments, the changing dynamic parameter is selected as the size of a geometry of the body. The change in the size of the body results in changes in the mass and inertia as they are automatically computed from the shape in the physics simulator. Table \ref{tab:dynamics} presents the details of the dynamics parameters used in the experiments for transfer to novel dynamics parameters.
\begin{table}[h!]
\centering
\begin{tabular}{c|c|c|c|c}
\textbf{Environment} & \textbf{Dynamics Parameter} $\mu$                   & \textbf{Range and Increments}        & \textbf{No. Samples} $p(\mu)$           & \textbf{Default of DMC} \\
\hline
\hline
\textbf{Cheetah}     & Torso length                              & $[0.3, 0.7]$, 0.01 increments & 41 samples  & 0.5                         \\
\textbf{Walker}      & Torso length                              & $[0.1, 0.5]$, 0.01 increments & 41 samples   & 0.3                         \\
\textbf{Finger}      & Finger length and distance to spinner & $[0.1, 0.4]$,  0.01 increments & 31 samples   & 0.16                       
\end{tabular}
\caption{Details of the dynamics parameters used in experiments for transfer to novel \textbf{dynamics} settings.}
\label{tab:dynamics}
\end{table}

\subsubsection{Reward and Dynamic Parameters.} Based on the reward and dynamics parameters described above, Table \ref{tab:rewards_dynamics} presents the details of the reward and dynamics parameters. Importantly, the space of parameters now forms a 2D grid.
\begin{table}[h!]
\centering
\begin{tabular}{c|c|c|c}
\textbf{Environment} & \textbf{Reward Parameter $\psi$} & \textbf{Dynamics Parameter $\mu$} & \textbf{Samples from $p(\psi, \mu)$} \\
\hline
\hline
\textbf{Cheetah}              & $[+1, +10]$, 1 increments                                 & $[0.3, 0.7]$, 0.05 increments              & $10 \times 9$ grid                   \\
\textbf{Walker}               & $[+1, +5]$, 0.5 increments                                & $[0.1, 0.5]$, 0.05 increments              & $10 \times 9$ grid                   \\
\textbf{Finger}               & $[+1, +15]$, 1 increments                                 & $[0.1, 0.4]$, 0.05 increments              & $15 \times 9$ grid                  
\end{tabular}
\caption{Details of the reward and dynamics parameters used in experiments for transfer to novel \textbf{rewards and dynamics} settings.}
\label{tab:rewards_dynamics}
\end{table}

\subsection{Data Collection}
\label{supp:data_collection}
First, the sampled MDPs $\M_i$ from each parameterized MDP family $\mathscr{M}$ is used to independently train a TD3 \cite{fujimoto2018addressing} agent for 1 million steps. Results of this phase are presented in Appendix \ref{supp:rl_solutions}. Consequently, each trained agent is rolled out for 10 episodes and the near-optimal trajectory is added to the dataset. The train ($\%85$) and test ($\%15$) MDPs are randomly selected from the set of all MDPs for 5 different seeds. 

\emph{The generated dataset will be released publicly by the time of publication, as a benchmark for zero-shot transfer learning to novel dynamics and reward settings. }

\clearpage
\section{Full Results}
\label{supp:results}
In this section, we provide additional empirical results, to supplement the results of Section 4.2.
\subsection{Additional Results}
\label{supp:additional_results}
\begin{figure}[h!]
    \centering
    \includegraphics[width=0.98\textwidth]{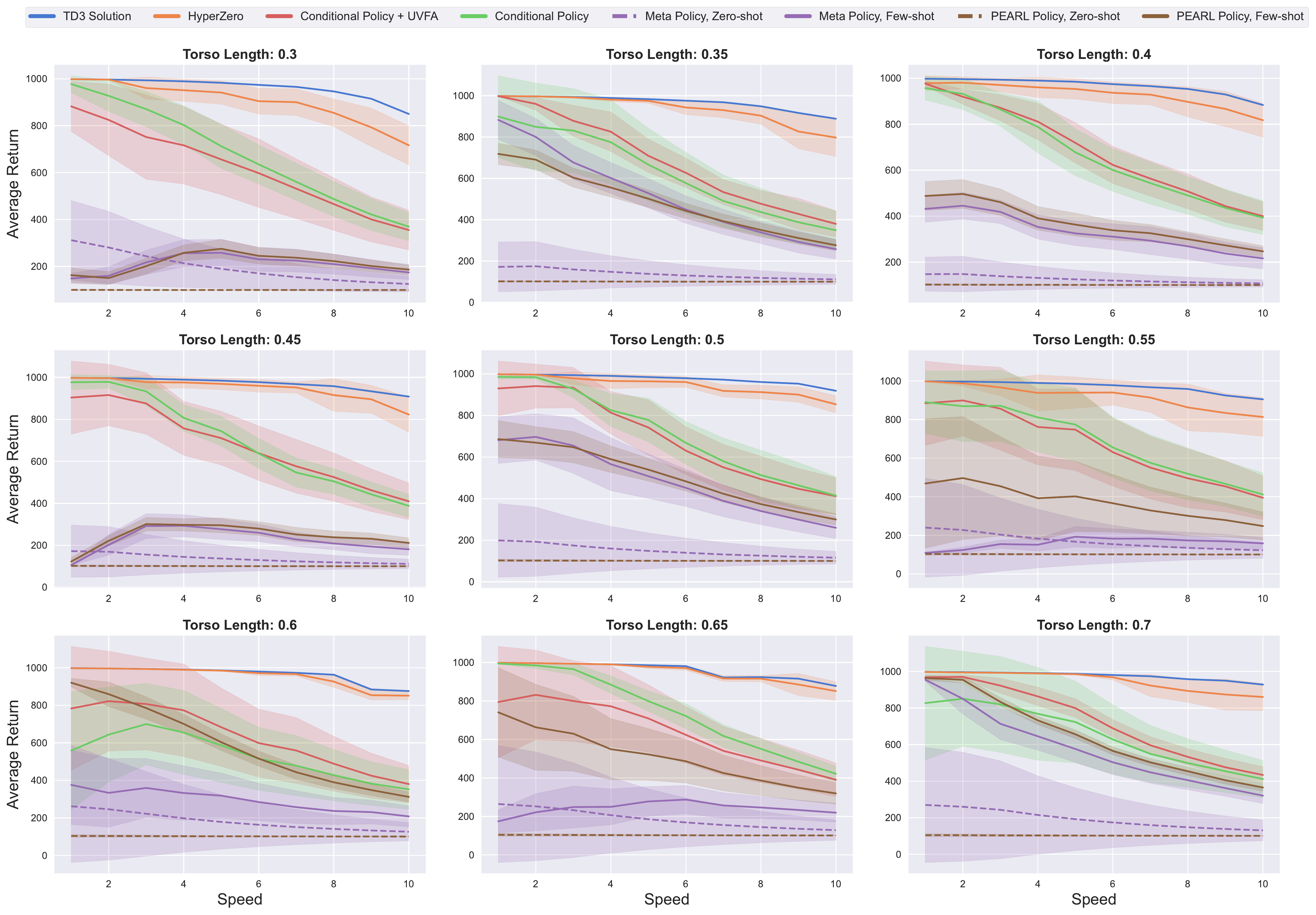}
    \caption{Zero-shot transfer to new \textbf{reward and dynamics} settings for the \textbf{Cheetah} environment, obtained on 5 seeds for random split of train/test tasks. Each subplot is for a specific value of the torso length which is a dynamics parameter $\mu_i$. This grid of 2D plots is to supplement the 3D plot of Figure 5.a of the paper. Solid lines present the mean and shaded regions present the standard deviation of the average return across the seeds. Horizontal axis shows the desired speed, which is a function of the reward parameters $\psi_i$.}
    \label{fig:rew_dyn_grid_cheetah}
\end{figure}

\begin{figure}[h!]
    \centering
    \includegraphics[width=0.98\textwidth]{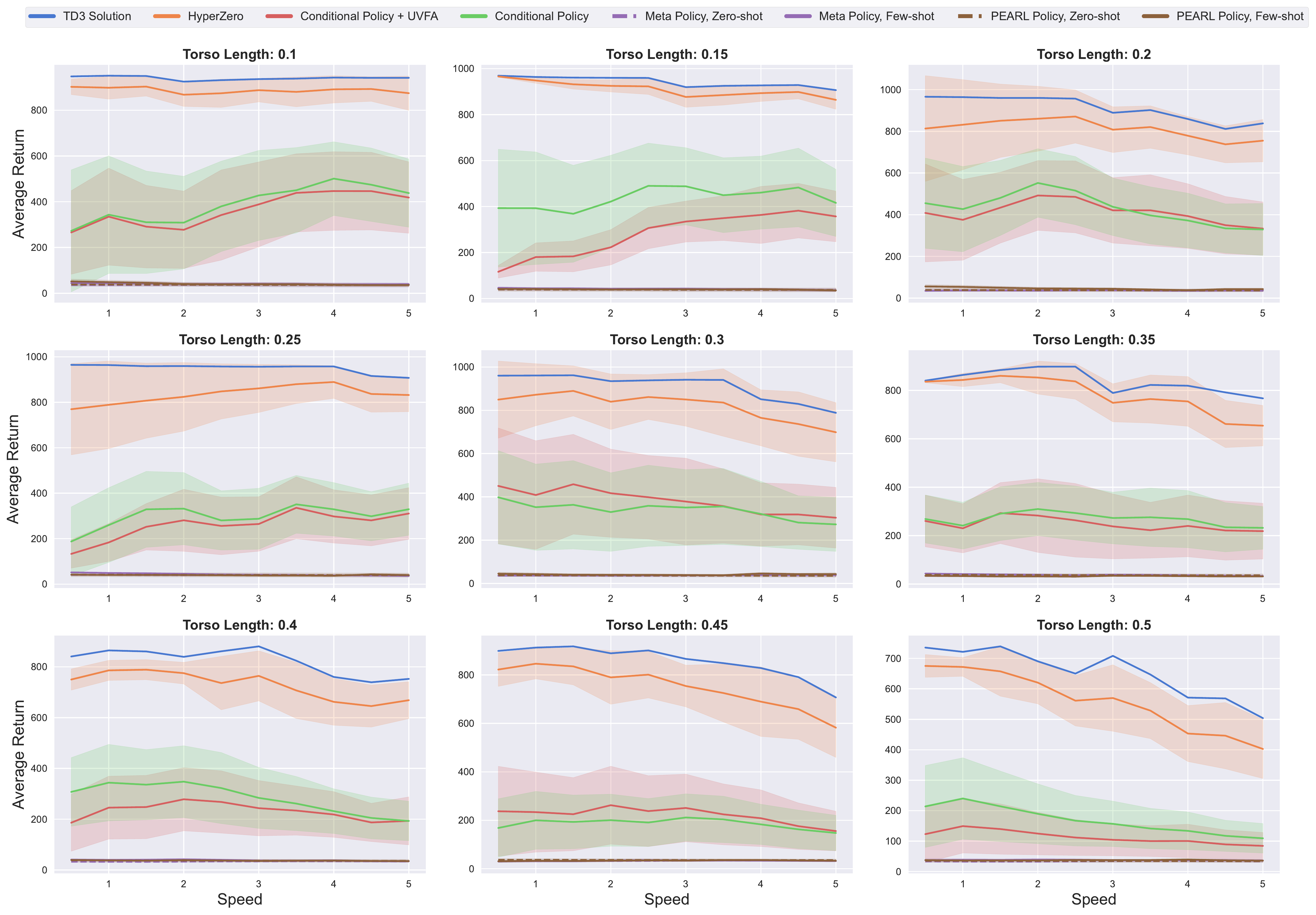}
    \caption{Zero-shot transfer to new \textbf{reward and dynamics} settings for the \textbf{Walker} environment, obtained on 5 seeds for random split of train/test tasks. Each subplot is for a specific value of the torso length which is a dynamics parameter $\mu_i$. This grid of 2D plots is to supplement the 3D plot of Figure 5.b of the paper. Solid lines present the mean and shaded regions present the standard deviation of the average return across the seeds. Horizontal axis shows the desired speed, which is a function of the reward parameters $\psi_i$.}
    \label{fig:rew_dyn_grid_walker}
\end{figure}

\begin{figure}[h!]
    \centering
    \includegraphics[width=0.98\textwidth]{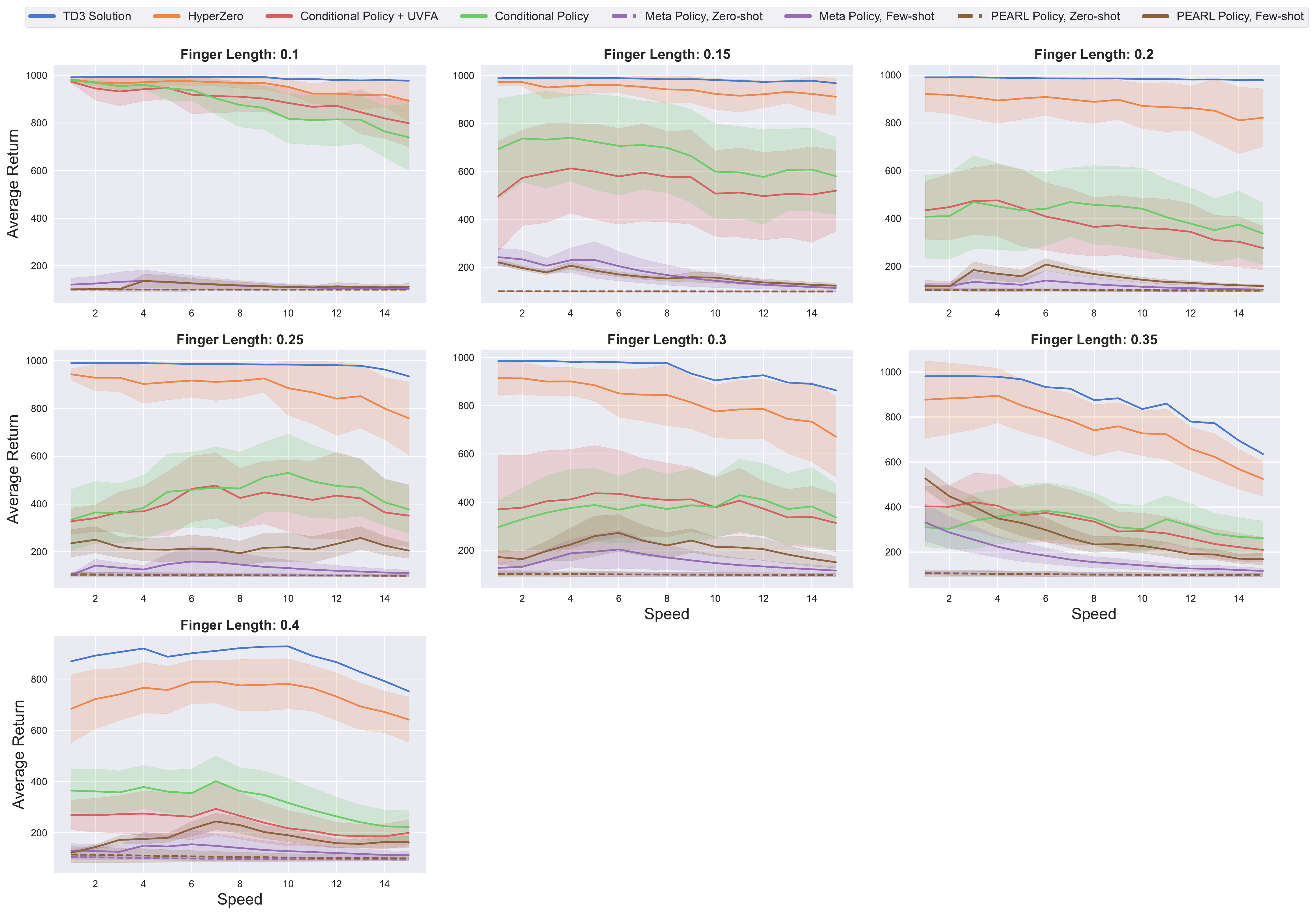}
    \caption{Zero-shot transfer to new \textbf{reward and dynamics} settings for the \textbf{Finger} environment, obtained on 5 seeds for random split of train/test tasks. Each subplot is for a specific value of the finger length which is a dynamics parameter $\mu_i$. This grid of 2D plots is to supplement the 3D plot of Figure 5.c of the paper. Solid lines present the mean and shaded regions present the standard deviation of the average return across the seeds. Horizontal axis shows the desired speed, which is a function of the reward parameters $\psi_i$.}
    \label{fig:rew_dyn_grid_finger}
\end{figure}

Figures \ref{fig:rew_dyn_grid_cheetah}-\ref{fig:rew_dyn_grid_finger} present grids of 2D plots of the experiments for zero-shot transfer to new rewards and dynamics settings. These Figures supplement 3D plots of Figure 5 of the paper as they provide a clearer comparison between HyperZero and the baselines. 

\clearpage
\subsection{Example Learned Behaviors}
\label{supp:example_behavior}
\begin{figure*}[h!]
    \centering
    \includegraphics[width=0.68\textwidth]{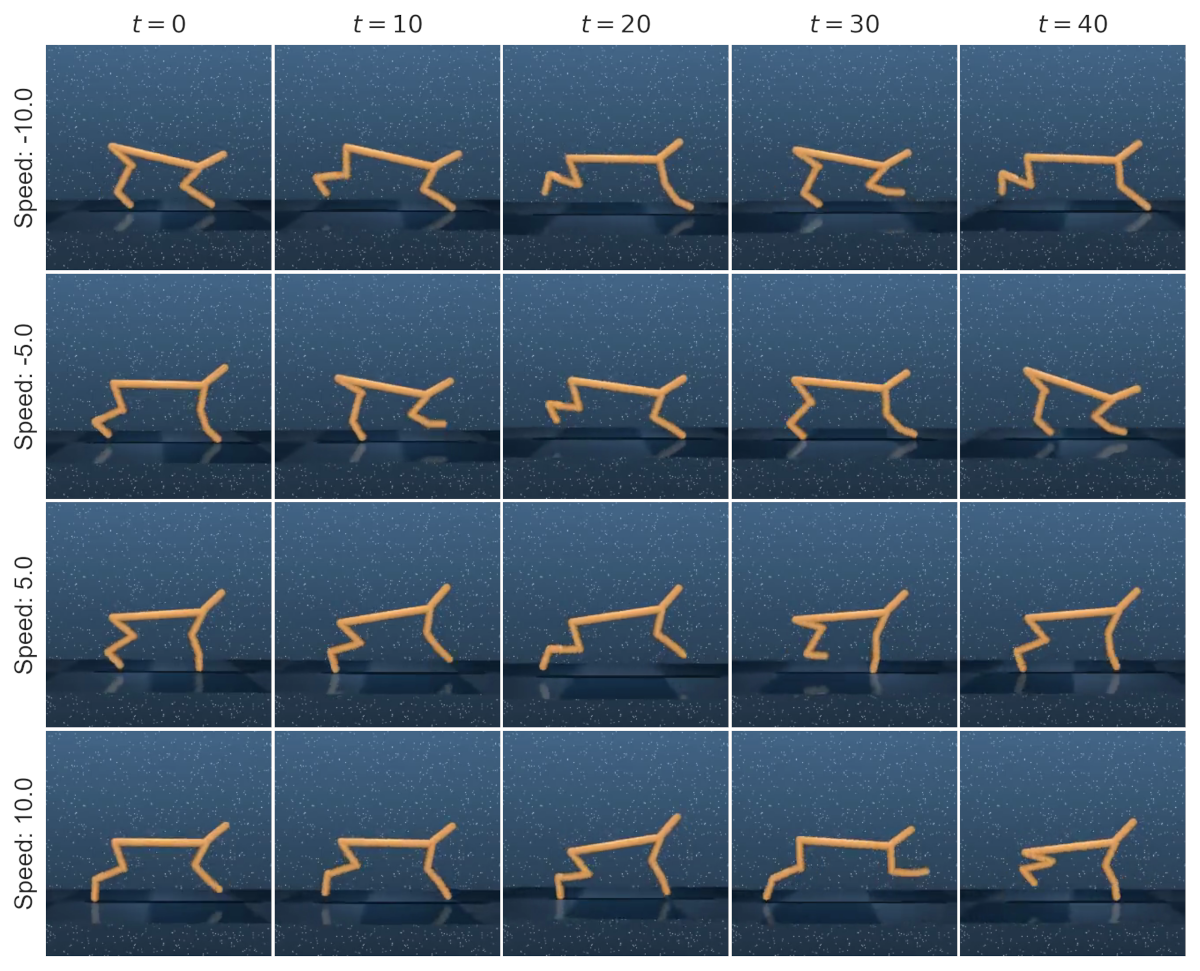}
    \caption{Example trajectories for the \textbf{Cheetah} environment with \textbf{reward} changes, obtained by rolling out a single HyperZero with different inputs.}
    \label{fig:example_rew_cheetah}
\end{figure*}

\begin{figure*}[h!]
    \centering
    \includegraphics[width=0.68\textwidth]{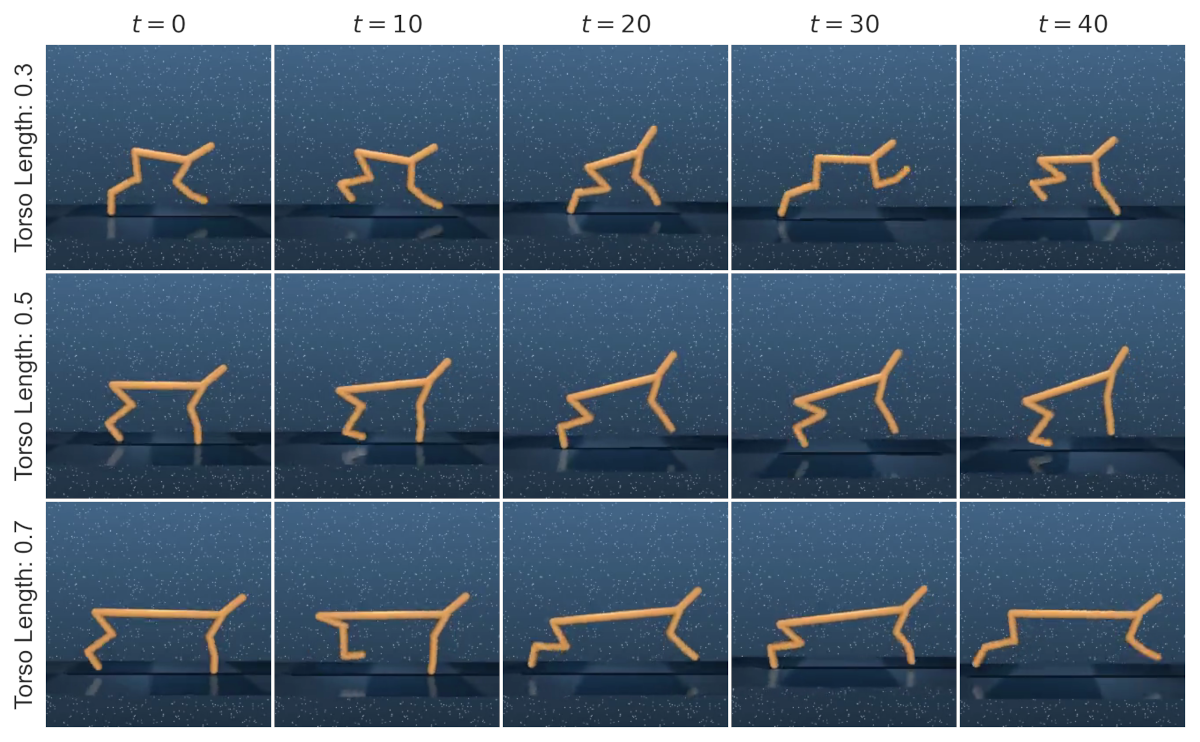}
    \caption{Example trajectories for the \textbf{Cheetah} environment with \textbf{dynamics} changes, obtained by obtained by rolling out a single HyperZero with different inputs.}
    \label{fig:example_dyn_cheetah}
\end{figure*}

\begin{figure*}[h!]
    \centering
    \includegraphics[width=0.68\textwidth]{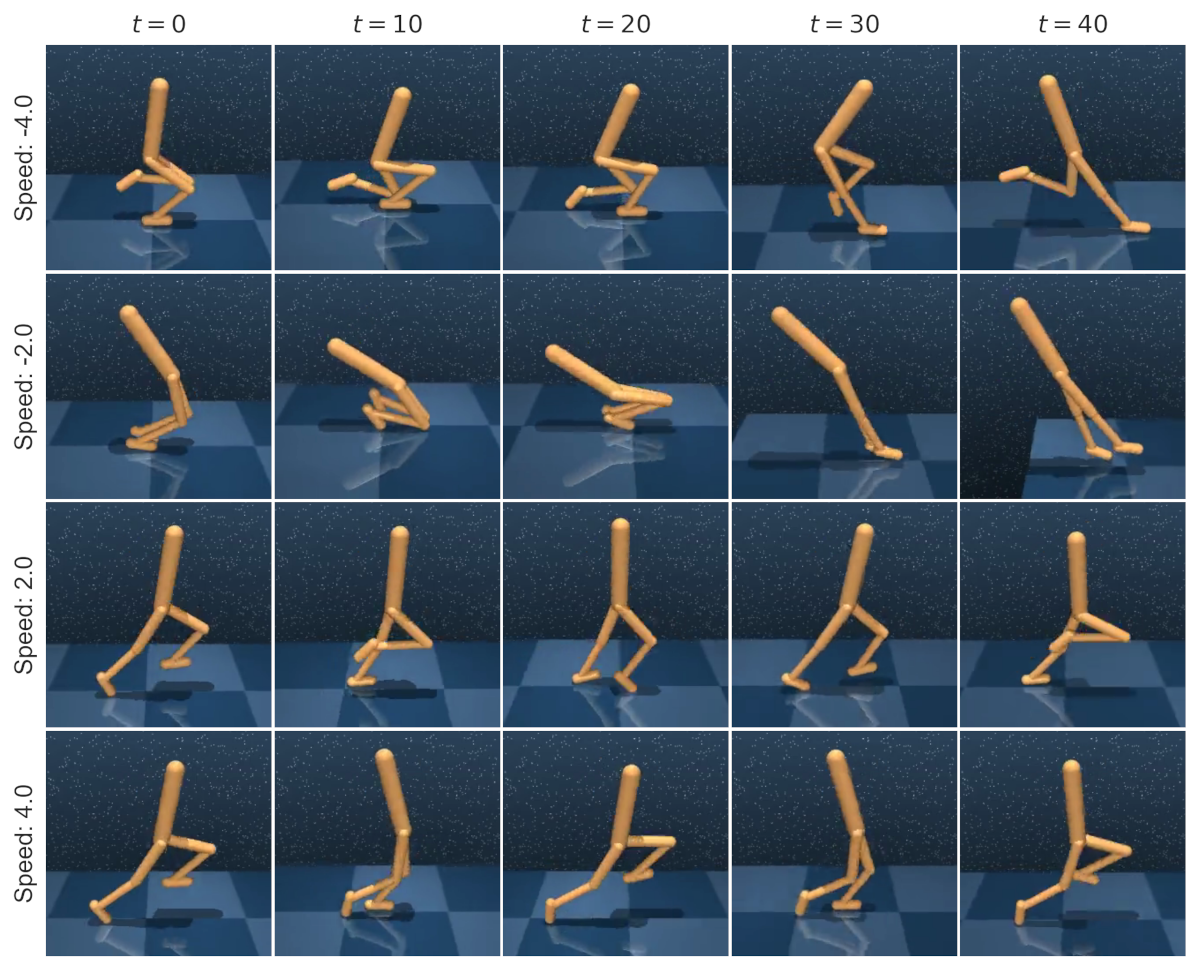}
    \caption{Example trajectories for the \textbf{Walker} environment with \textbf{reward} changes, obtained by rolling out a single HyperZero with different inputs.}
    \label{fig:example_rew_walker}
\end{figure*}

\begin{figure*}[h!]
    \centering
    \includegraphics[width=0.68\textwidth]{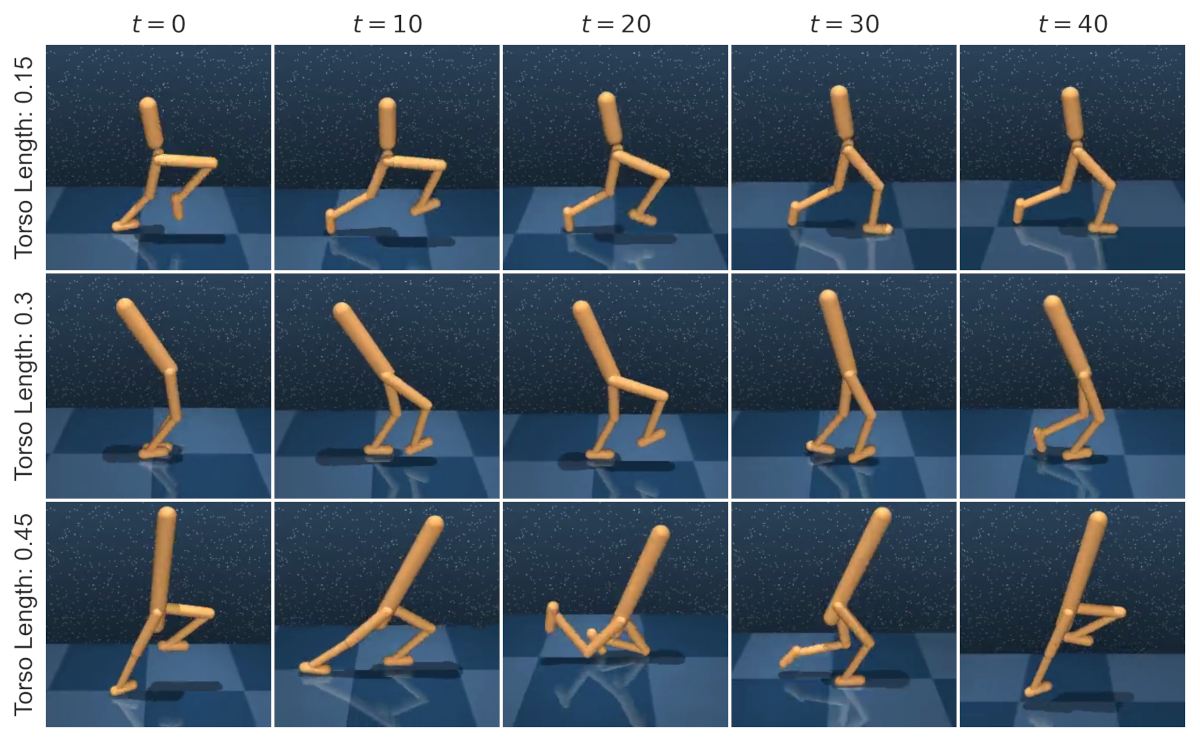}
    \caption{Example trajectories for the \textbf{Walker} environment with \textbf{dynamics} changes, obtained by rolling out a single HyperZero with different inputs.}
    \label{fig:example_dyn_walker}
\end{figure*}

\begin{figure*}[h!]
    \centering
    \includegraphics[width=0.68\textwidth]{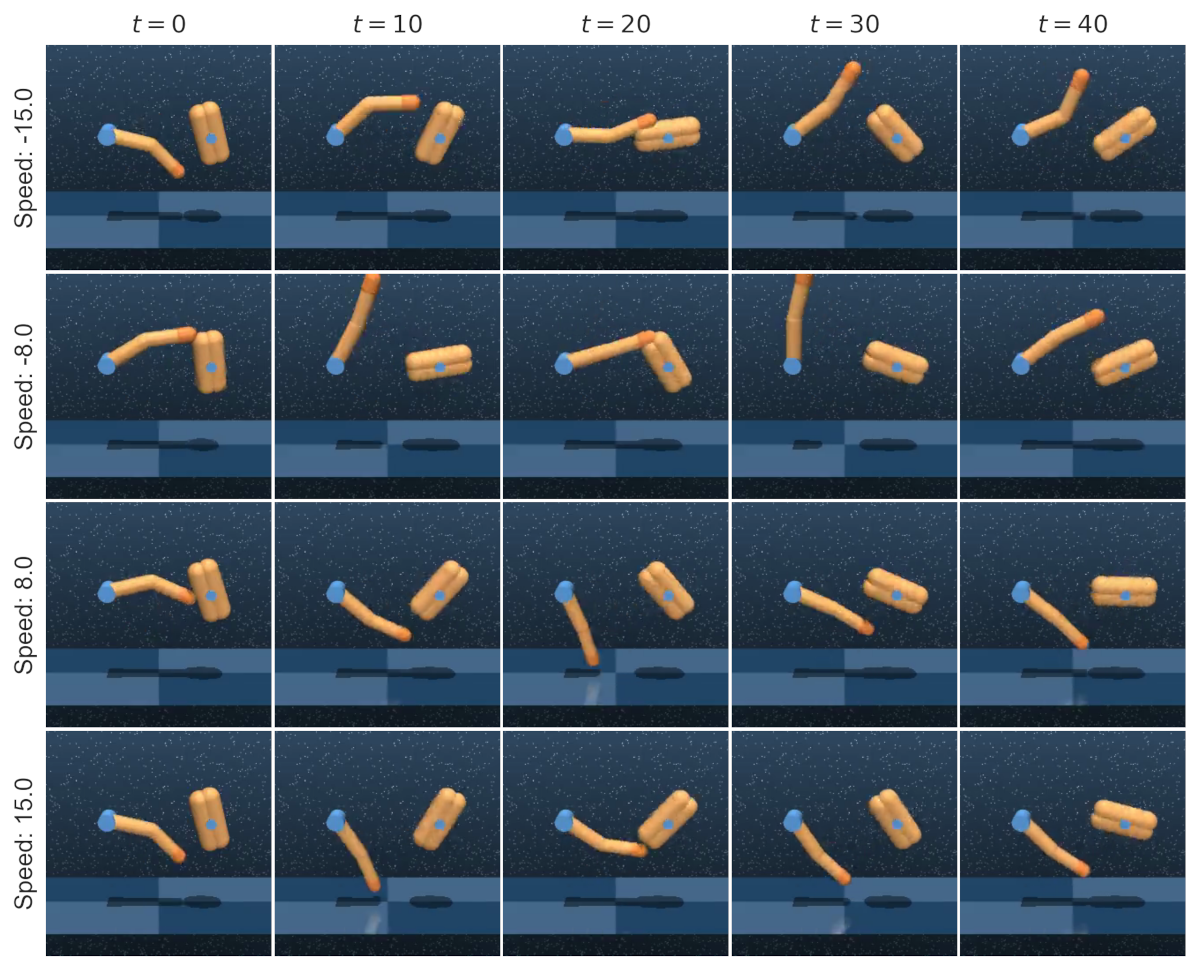}
    \caption{Example trajectories for the \textbf{Finger} environment with \textbf{reward} changes, obtained by rolling out a single HyperZero with different inputs.}
    \label{fig:example_rew_finger}
\end{figure*}

\begin{figure*}[h!]
    \centering
    \includegraphics[width=0.68\textwidth]{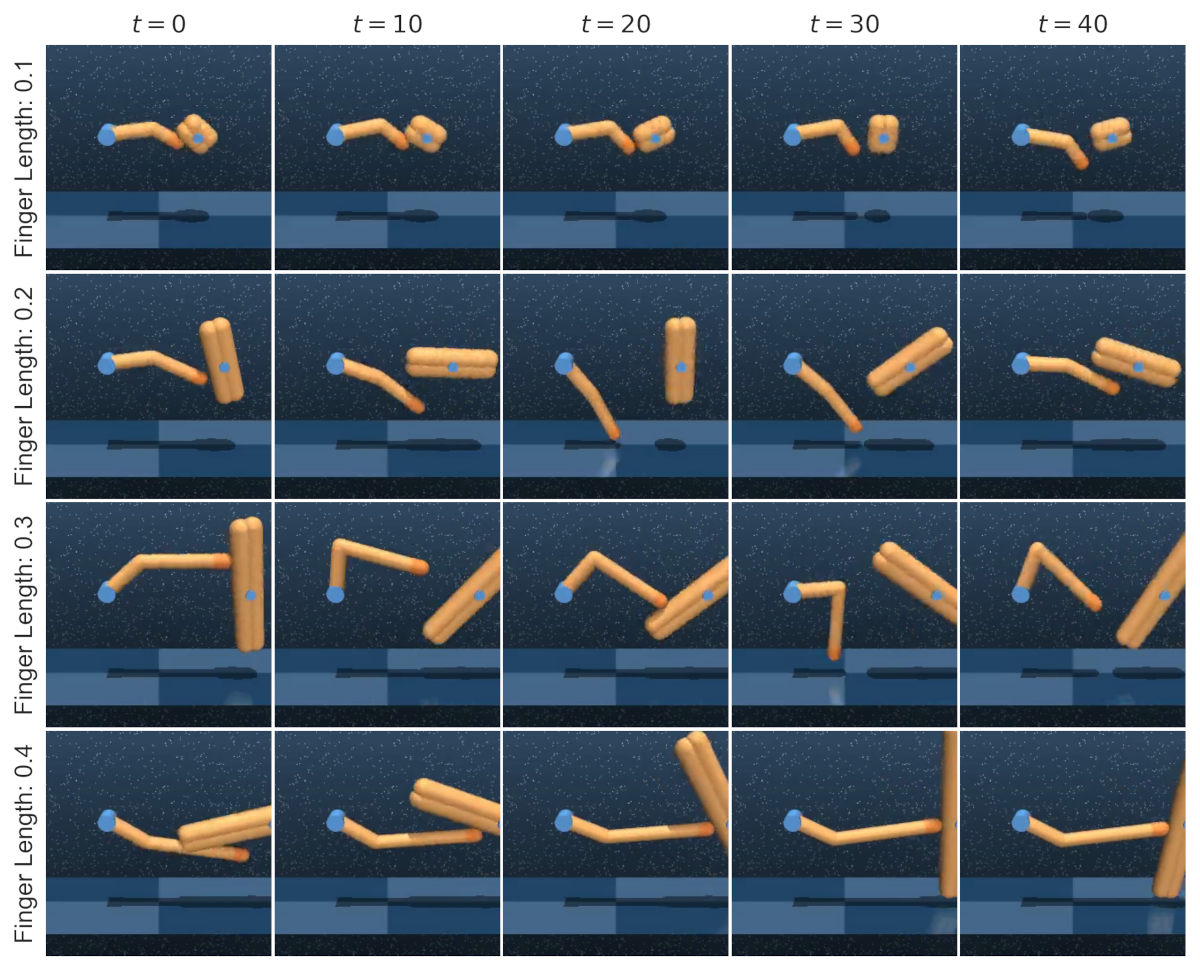}
    \caption{Example trajectories for the \textbf{Finger} environment with \textbf{dynamics} changes, obtained by rolling out a single HyperZero with different inputs.}
    \label{fig:example_dyn_finger}
\end{figure*}

Figures \ref{fig:example_rew_cheetah}-\ref{fig:example_dyn_finger} show example trajectories obtained by rolling out the trained HyperZero policy on different reward and dynamics settings for various environments. In each case, the trajectories are generated by rolling out a single HyperZero agent with different inputs, corresponding to the MDP setting. \emph{Please refer to the supplementary video submission for videos of these behaviours.}




\clearpage
\subsection{Training of the Optimal RL Solution}
\label{supp:rl_solutions}
In this section, we present the learning curves for the actual RL solutions that were used as the near-optimal solutions to each parameterized MDP $\M_i \in \mathscr{M}$. As described in Appendix \ref{supp:data_collection}, the near-optimal policy $\pi^*_i$ and action-value function $Q^*_i$ for each MDP $\M_i \in \mathscr{M}$ is obtained by training TD3 \cite{fujimoto2018addressing} for 1 million steps. 

\begin{figure*}[h!]
    \centering
    \begin{subfigure}[b]{0.33\textwidth}
        \includegraphics[width=\textwidth]{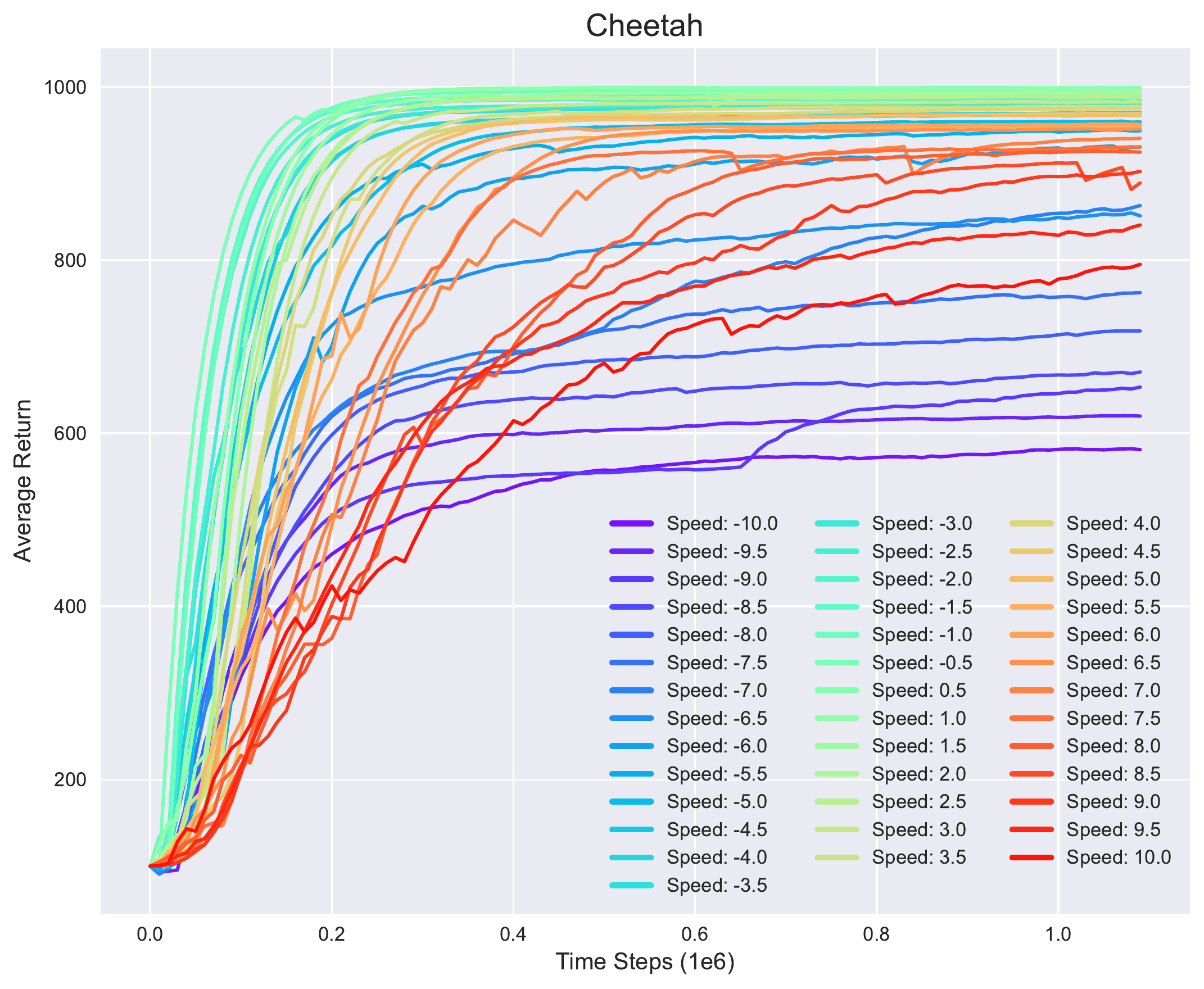}
        \caption{Cheetah environment.}
    \end{subfigure}
    \hfill
    \begin{subfigure}[b]{0.33\textwidth}
        \includegraphics[width=\textwidth]{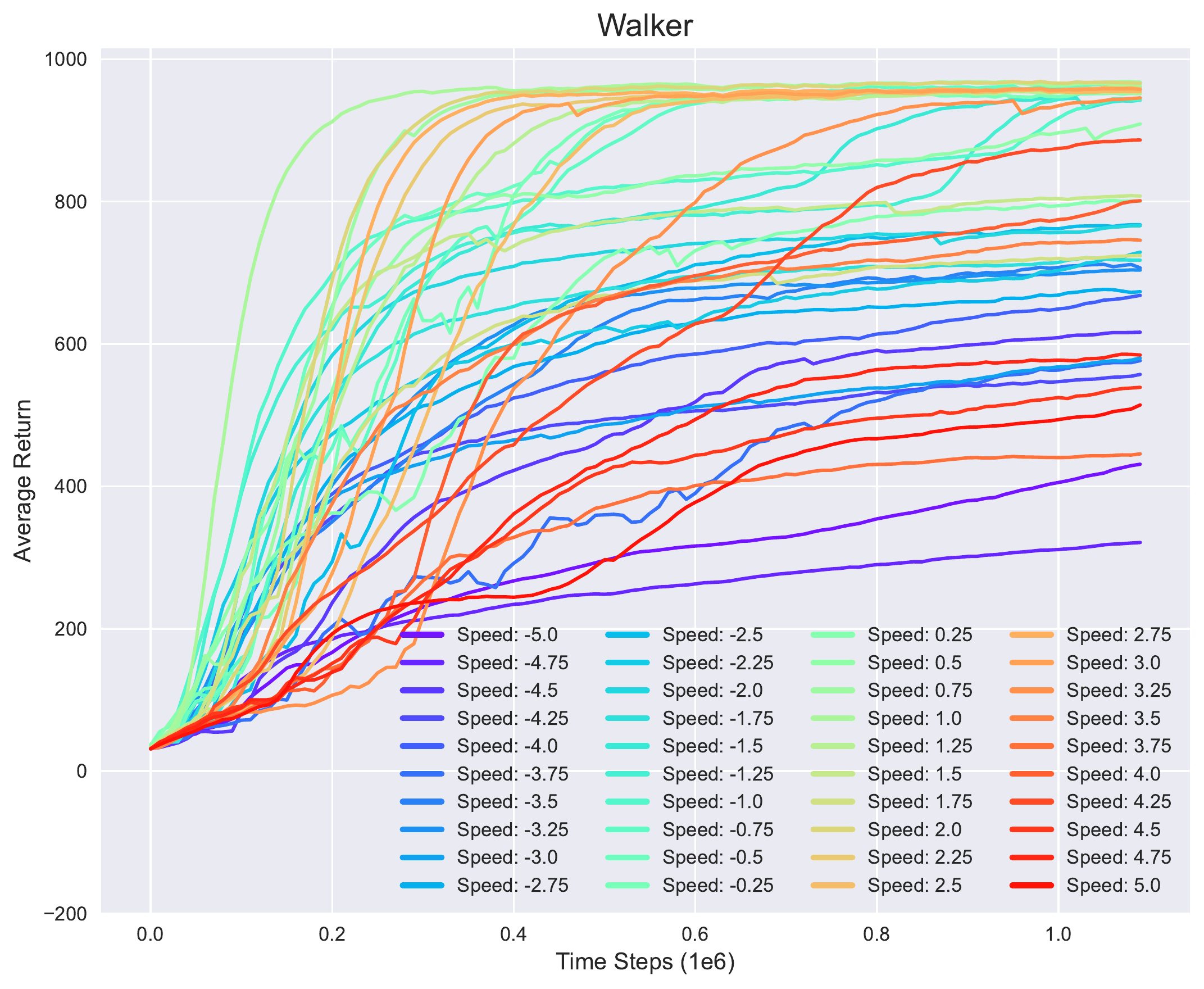}
        \caption{Walker environment.}
    \end{subfigure}
    \hfill
    \begin{subfigure}[b]{0.33\textwidth}
        \includegraphics[width=\textwidth]{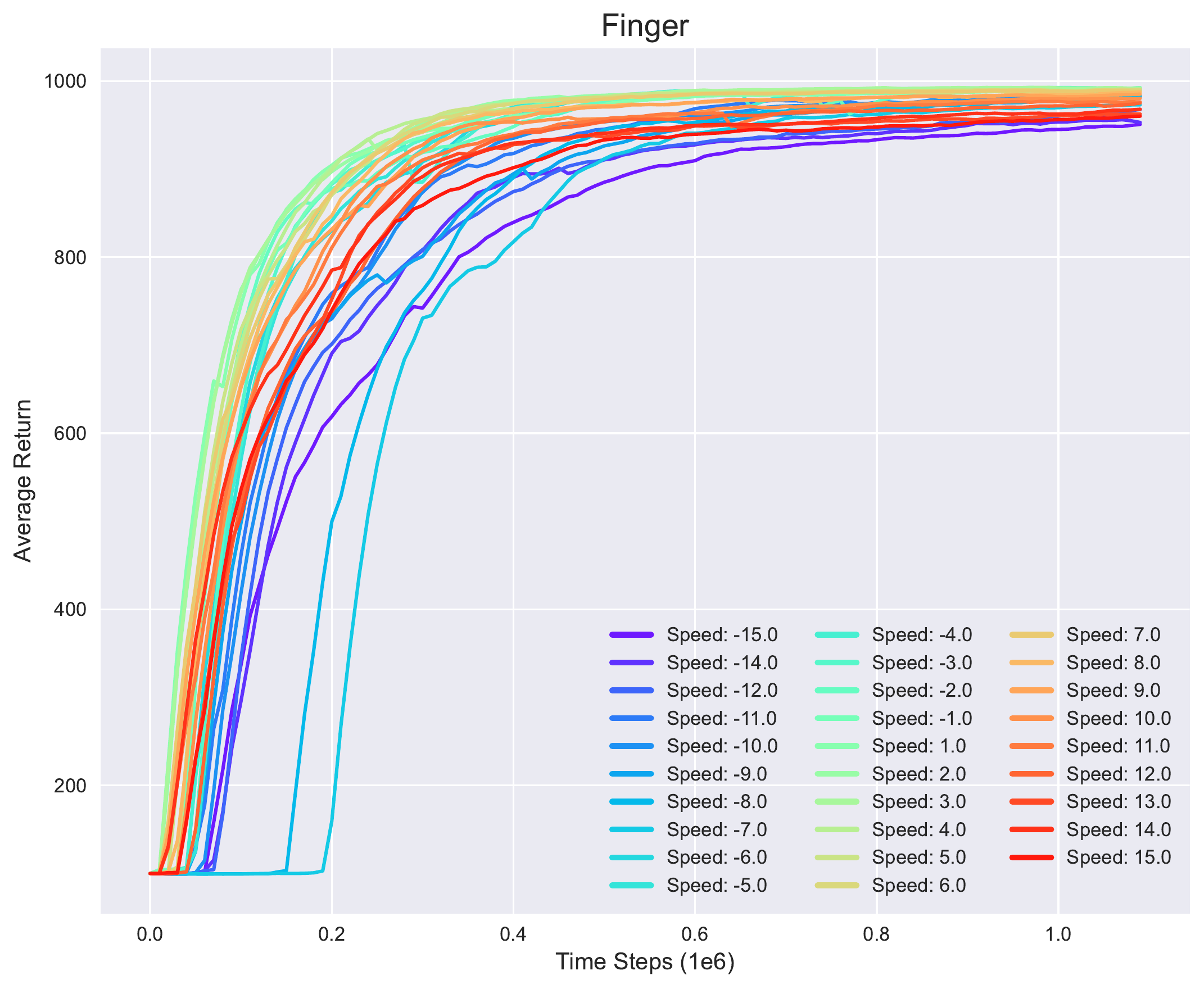}
        \caption{Finger environment.}
    \end{subfigure}
    \caption{Learning curves for TD3 obtained on environments with \textbf{reward changes}.  \textbf{(a)} The desired speed of the cheetah is changed from -10 to +10 with 0.5 increments while its torso length is set to the default value in DM control (0.5). \textbf{(b)} The desired speed of the walker is changed from -5 to +5 with 0.25 increments while its torso length is set to the default value in DM control (0.1). \textbf{(c)} The desired spinning speed is changed from -15 to +15 with 1 increments while the finger length is set to the default value in DM control (0.16)}
    \label{fig:rl_solutions_rew}
\end{figure*}

\begin{figure*}[h!]
    \centering
    \begin{subfigure}[b]{0.33\textwidth}
        \includegraphics[width=\textwidth]{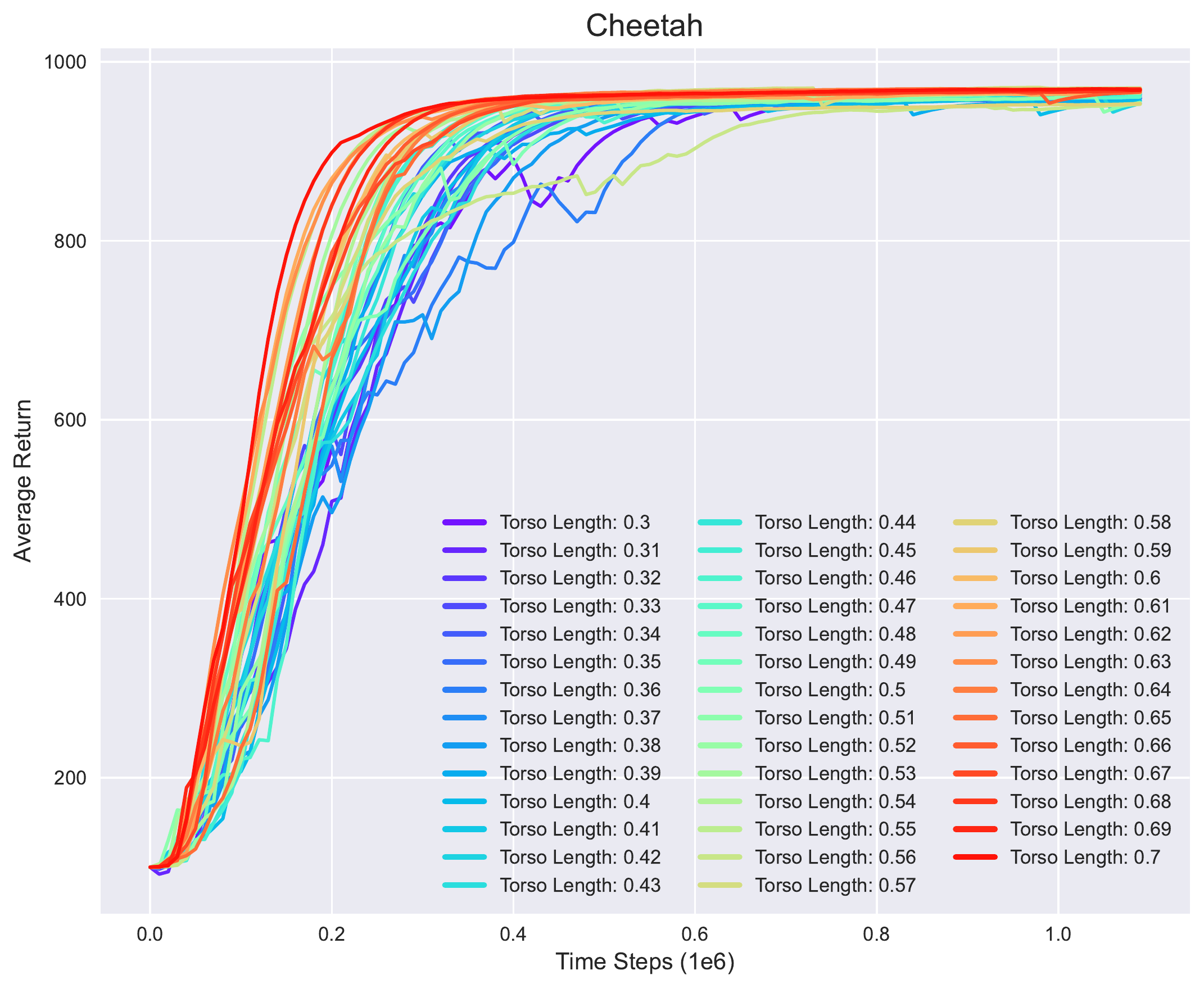}
        \caption{Cheetah environment.}
    \end{subfigure}
    \hfill
    \begin{subfigure}[b]{0.33\textwidth}
        \includegraphics[width=\textwidth]{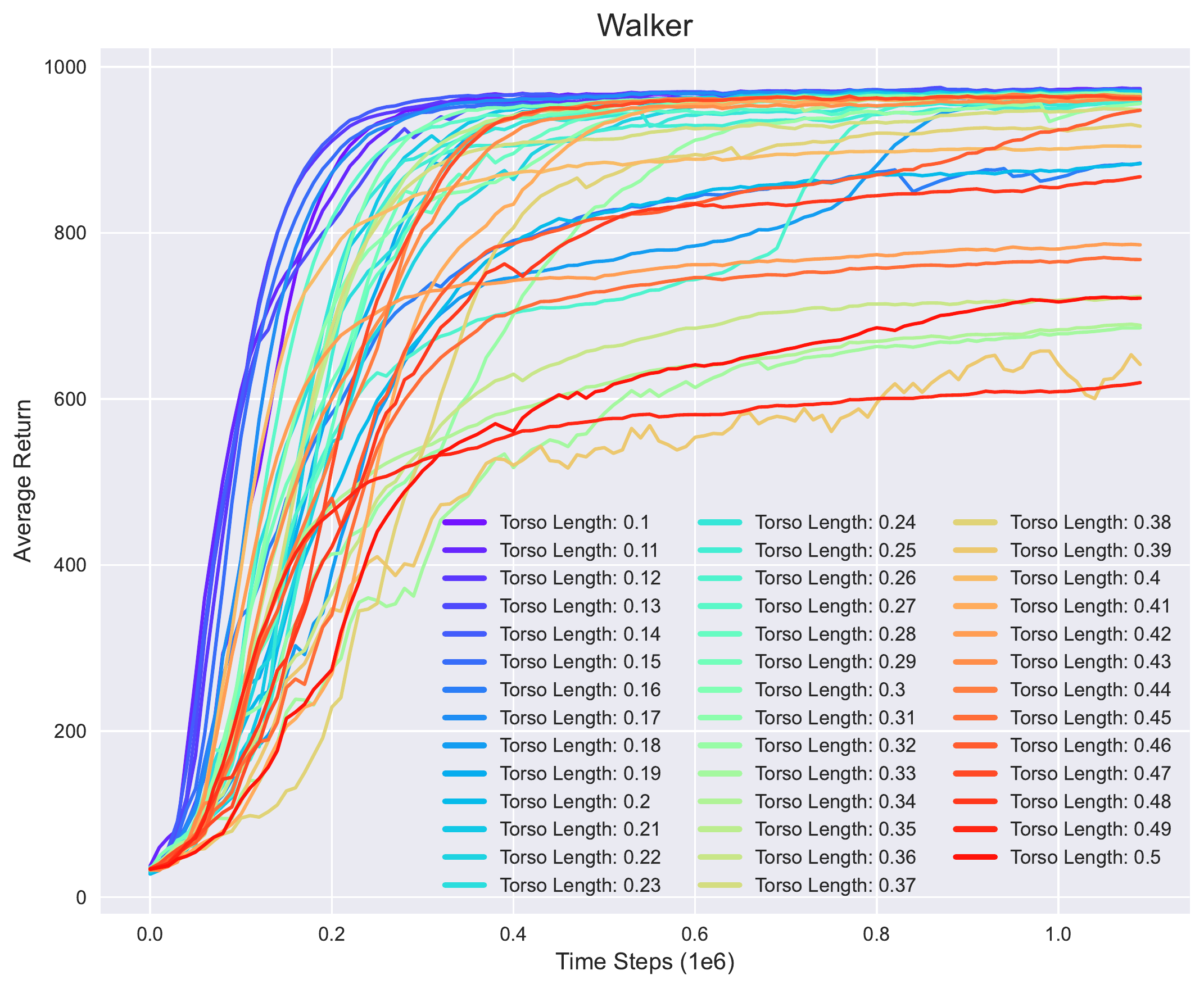}
        \caption{Walker environment.}
    \end{subfigure}
    \hfill
    \begin{subfigure}[b]{0.33\textwidth}
        \includegraphics[width=\textwidth]{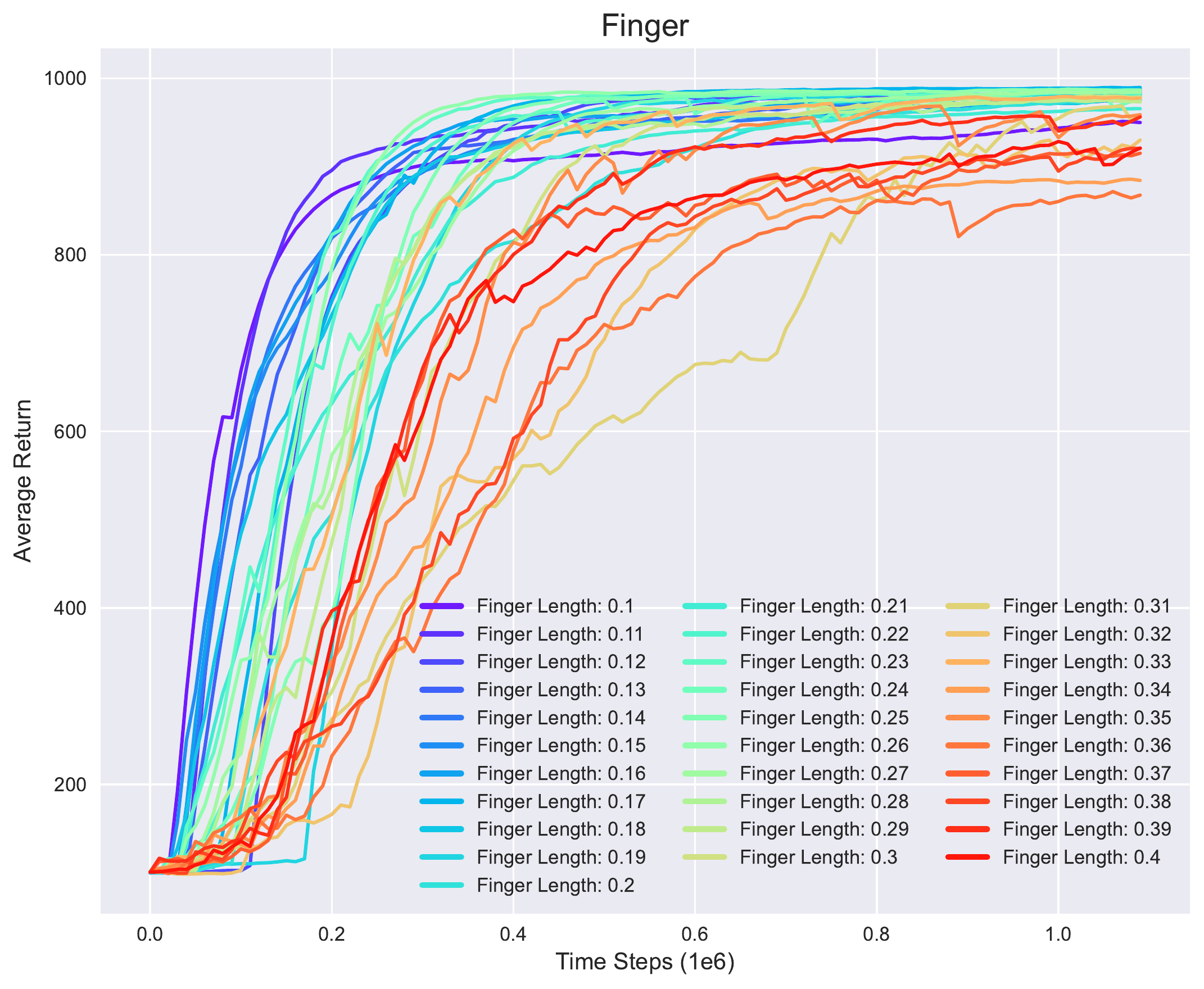}
        \caption{Finger environment.}
    \end{subfigure}
    \caption{Learning curves for TD3 obtained on environments with \textbf{dynamics changes}.  \textbf{(a)} The torso length of the cheetah is changed from 0.3 to 0.7 with 0.01 increments while its desired speed is set to slow running (+5). \textbf{(b)} The torso length of the walker is changed from 0.1 to 0.5 with 0.01 increments while its desired speed is set to  walking (+1). \textbf{(c)} The finger length is changed from 0.1 to 0.4 with 0.01 increments while the desired spinning speed is set to +15.}
    \label{fig:rl_solutions_dyn}
\end{figure*}

\begin{figure*}[h!]
    \centering
    \begin{subfigure}[b]{0.33\textwidth}
        \includegraphics[width=\textwidth]{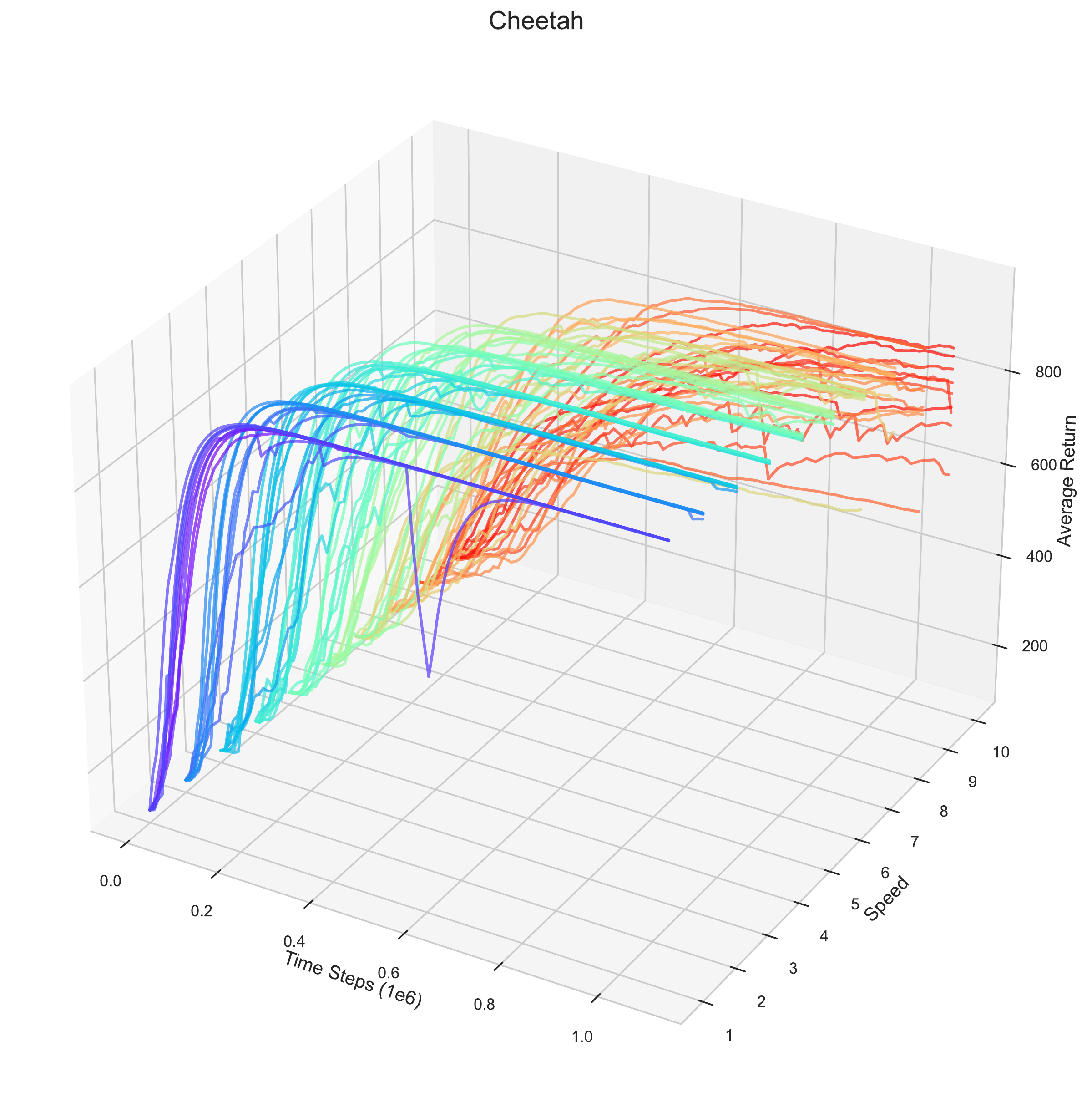}
        \caption{Cheetah environment.}
    \end{subfigure}
    \hfill
    \begin{subfigure}[b]{0.33\textwidth}
        \includegraphics[width=\textwidth]{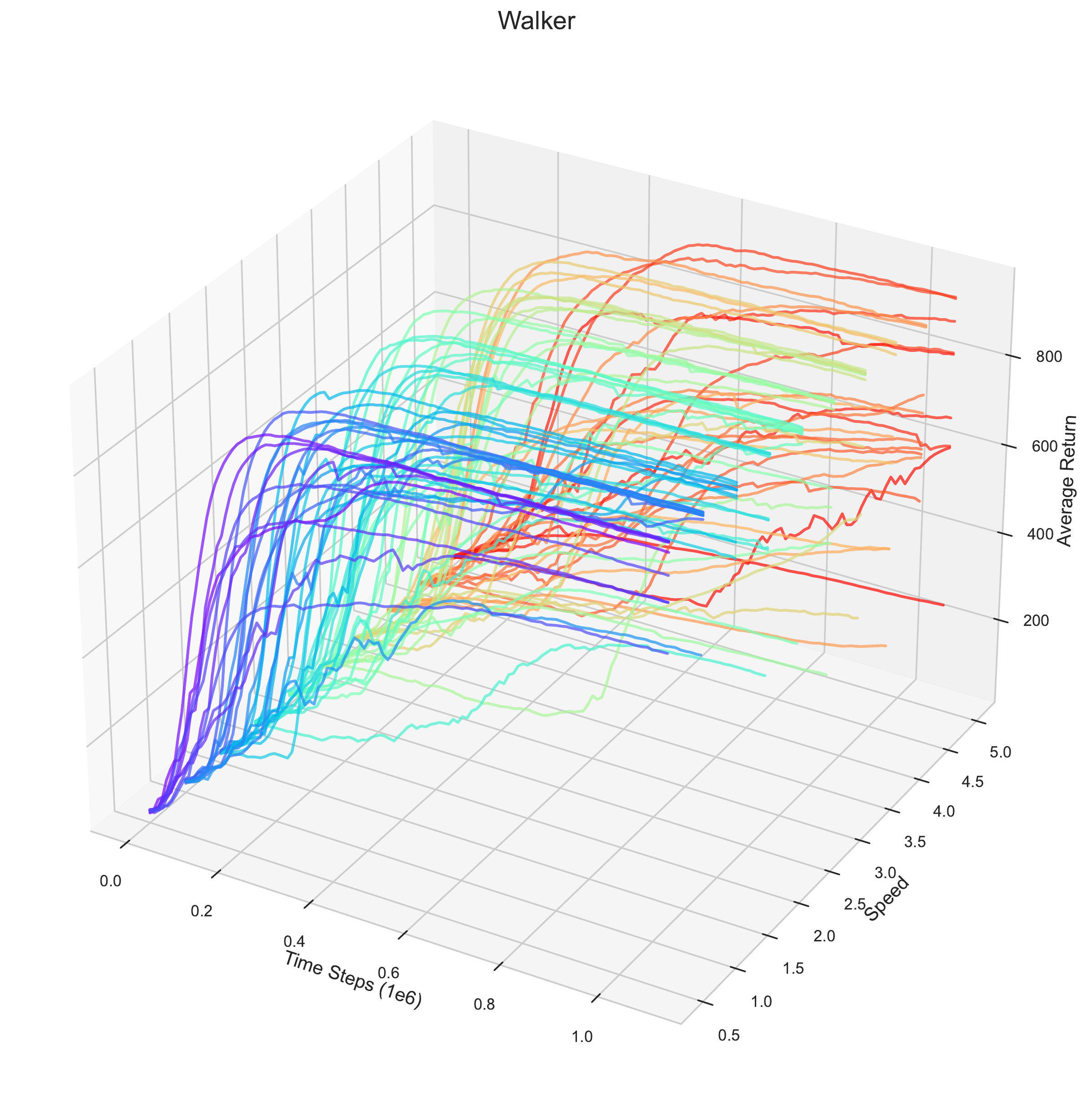}
        \caption{Walker environment.}
    \end{subfigure}
    \hfill
    \begin{subfigure}[b]{0.33\textwidth}
        \includegraphics[width=\textwidth]{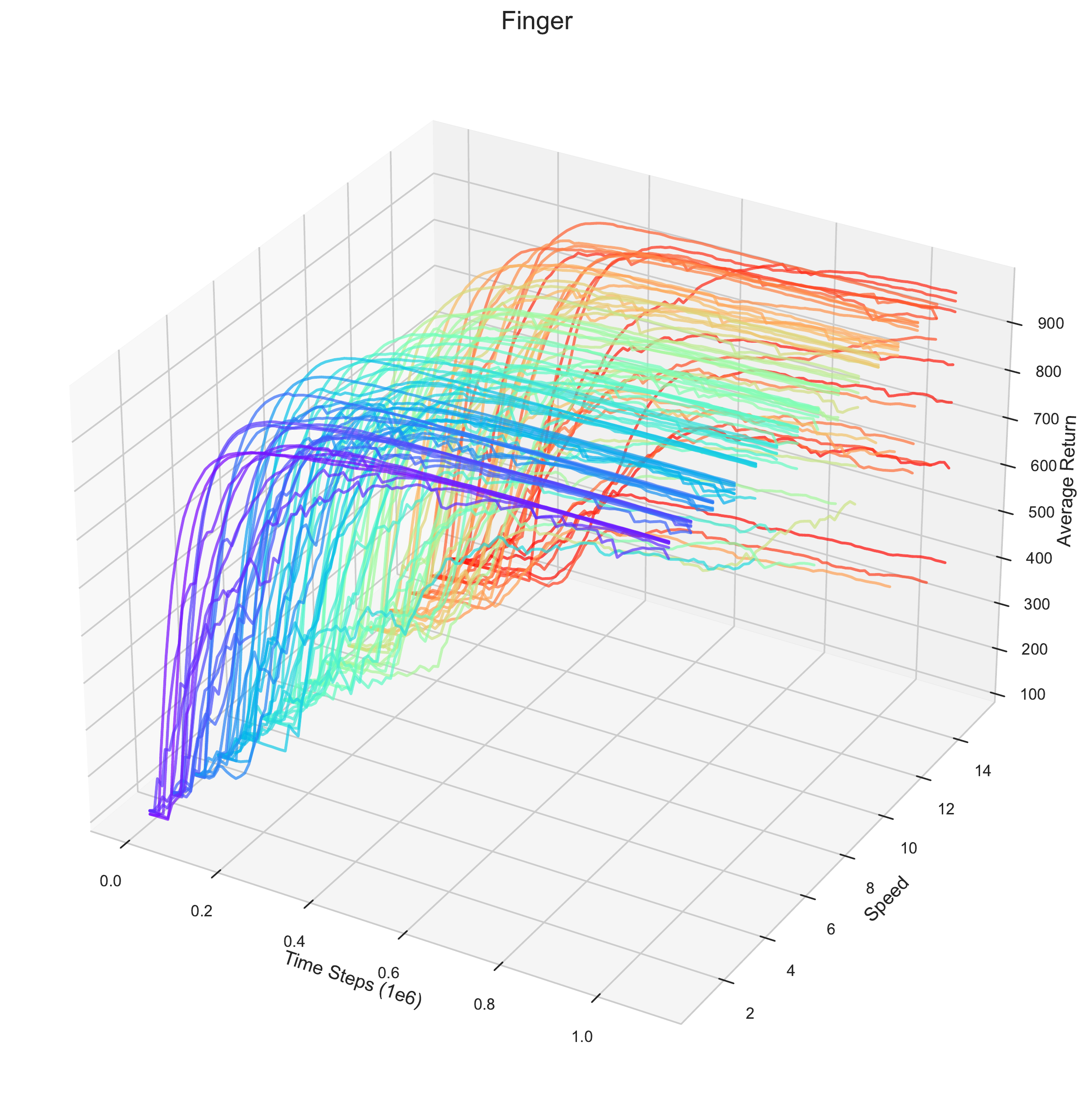}
        \caption{Finger environment.}
    \end{subfigure}
    \caption{Learning curves for TD3 obtained on environments with \textbf{dynamics and rewards changes}. Different colors correspond to different dynamics parameters. \textbf{(a)} The torso length of the cheetah is changed from 0.3 to 0.7 with 0.05 increments while its desired speed is changed from +1 to +10 with 1 increments. \textbf{(b)} The torso length of the walker is changed from 0.1 to 0.5 with 0.05 increments while its desired speed is changed from +0.5 to +5. \textbf{(c)} The finger length is changed from 0.1 to 0.4 with 0.05 increments while its desired speed is changed from +1 to +15.}
    \label{fig:rl_solutions_rew_dyn}
\end{figure*}

Figures \ref{fig:rl_solutions_rew} and \ref{fig:rl_solutions_dyn} show the results obtained for individual changes in the reward and dynamics settings, respectively. Figure \ref{fig:rl_solutions_rew_dyn} shows the results obtained for the simultaneous changes in the reward and dynamics settings. 

As these results suggest, in some instances the TD3 solver appears to be not fully-converged to the near-optimal solution. Therefore, in practice our method works despite the violation of Assumption 2.

\clearpage
\section{Implementation Details}
\label{supp:hyperparam_details}
\emph{Our HyperZero implementation, as well as the full learning pipeline for zero-shot transfer learning by approximating RL solutions, will be made publicly available by the time of publication.} 

We implemented our method in PyTorch \cite{paszke2017automatic} and results were obtained using Python v3.9.12, PyTorch 1.10.1, CUDA 11.1, and Mujoco 2.1.1 \cite{todorov2012mujoco} on Nvidia RTX A6000 GPUs.

\subsubsection{RL Solver.} We use TD3 \cite{fujimoto2018addressing} for obtaining the near-optimal RL solutions with the hyperparamters of Table \ref{tab:hyperparams_td3}. 

\begin{table}[h!]
\centering
\begin{tabular}{cc}
\hline
\textbf{Hyperparameter}                       & \textbf{Setting}         \\
\hline
Learning rate                            & $1$e$-4$                   \\
Optimizer                                & Adam \cite{kingma2014adam}                     \\
Mini-batch size                          & $256$                      \\
Actor update frequency $d$               & $2$                        \\
Target networks update frequency         & $2$                        \\
Target networks soft-update $\tau$       & $0.01$                     \\
Target policy smoothing stddev. clip $c$ & 0.3                      \\
Hidden dim.                              & $256$                      \\
Replay buffer capacity                   & $10^6$                   \\
Discount $\gamma$                        & $0.99$                     \\
Seed frames                              & $4000$                     \\
Exploration steps                        & $2000$                     \\
Exploration stddev. schedule             & linear$(1.0, 0.1, 1$e$6)$ \\
\hline
\end{tabular}
\caption{TD3 hyperparameters for obtaining the near-optimal solution for each $\M_i \in \mathscr{M}$.}
\label{tab:hyperparams_td3}
\end{table}

\subsubsection{HyperZero.} HyperZero generates the weights of a near-optimal policy $\hat{\pi}^*_\theta$ and action-value function $\hat{Q}^*_\phi$ conditioned on the MDP $\M_i$ parameters. The MDP parameters are used as inputs to a shared task embedding, with an architecture based on MLP ResNet blocks \cite{he2016deep}, as shown in Listing \ref{lst:embedding}. The 256-dimensional output $z$ of the embedding is transformed linearly to generate the weights of each layer of the main networks, in a similar fashion as standard hypernetworks \cite{ha2016hypernetworks}. 
\begin{lstlisting}[caption={Architecture of the shared task embedding, used in HyperZero and all of the baselines.}, label={lst:embedding}, language=Python]
task_embedding = nn.Sequential(
    nn.Linear(mdp_param_dim, 64),
    ResBlock(64),
    ResBlock(64),

    nn.Linear(64, 128),
    ResBlock(128),
    ResBlock(128),

    nn.Linear(128, 256),
    ResBlock(256),
    ResBlock(256)
)

class ResBlock(nn.Module):
    def __init__(self, in_size):
        super().__init__()
        self.fc = nn.Sequential(
            nn.ReLU(),
            nn.Linear(in_size, in_size),
            nn.ReLU(),
            nn.Linear(in_size, in_size),
        )

    def forward(self, x):
        h = self.fc(x)
        return x + h
\end{lstlisting}

The generated policy and action-value functions are MLP networks with one hidden layer of 256 dimensions. The rest of the hyperparameters are detailed in Table \ref{tab:hyperparams_all}.

\begin{table}[h!]
\centering
\begin{tabular}{cc}
\hline
\textbf{Hyperparameter}                       & \textbf{Setting}         \\
\hline
Learning rate                            & $1$e$-4$                   \\
Optimizer                                & Adam \cite{kingma2014adam}                     \\
Mini-batch size                          & $512$                      \\
Hidden dim. of $\hat{\pi}^*_\theta$                             & $256$                      \\
Hidden dim. of $\hat{Q}^*_\phi$                             & $256$                      \\
Task embedding dim.  & $256$ \\
\hline
\end{tabular}
\caption{Hyperparameters of HyperZero and all of the baselines.}
\label{tab:hyperparams_all}
\end{table}

\subsubsection{Baselines.} For a fair comparison with HyperZero, all baselines use the same task embedding of Listing \ref{lst:embedding}. The output of this embedding is concatenated with the inputs to the policy (or value function) and all networks are trained simultaneously using the supervised learning loss functions described in Section 3. Similarly to HyperZero, the baselines use a 256-dimensional task embedding and an MLP with one hidden layer of 256 dimensions for the policy (or the value function).

Our MAML \cite{finn2017model} and PEARL \cite{rakelly2019efficient} implementations are based on the learn2learn package \cite{arnold2020learn2learn}. Since the MAML policy is context-conditioned, it can be regarded as an adaptation of PEARL \cite{rakelly2019efficient} in which the inferred task is substituted by the ground-truth task.  Hyperparameters specific to MAML and PEARL are presented in Table \ref{tab:hyperparams_maml}.

\begin{table}[h!]
\centering
\begin{tabular}{cc}
\hline
\textbf{Hyperparameter}                       & \textbf{Setting}         \\
\hline
Meta learning rate                            & $1$e$-4$                   \\
Fast learning rate                            & $1$e$-2$                   \\
Meta-batch size                          & $32$, or the total number of meta-train MDPs                      \\
$K$-shot                          & $10$                      \\
\hline
\end{tabular}
\caption{Hyperparameters specifc to MAML and PEARL.}
\label{tab:hyperparams_maml}
\end{table}

\end{document}